\PassOptionsToPackage{table}{xcolor}
\documentclass[10pt,twocolumn,letterpaper]{article}

\usepackage{cvpr}              % To produce the CAMERA-READY version
\usepackage{enumitem}
\usepackage{url}
\usepackage[utf8]{inputenc} % allow utf-8 input
\usepackage[T1]{fontenc}    % use 8-bit T1 fonts
\usepackage{booktabs}       % professional-quality 
\usepackage{amsfonts}       % blackboard math symbols
\usepackage{amsmath}
\usepackage{amssymb}
\usepackage{balance} % To balance columns in the references section
\definecolor{LightCyan}{rgb}{0.88,0.9,1}
\usepackage{nicefrac}       % compact symbols for 1/2, etc.
\usepackage{microtype}      % microtypography
\usepackage[table]{xcolor}  % colors
\usepackage{enumitem}
\usepackage{graphicx}
\usepackage{pifont}     % For \ding symbols
\newcommand{\cmark}{\ding{51}}%
\newcommand{\xmark}{\ding{55}}%
\usepackage{array}

\usepackage{times}
\usepackage{epsfig}
\usepackage{graphicx}
\usepackage{caption}
\usepackage[super]{nth}
\usepackage{amsmath}
\definecolor{cvprblue}{rgb}{0.21,0.49,0.74}
\usepackage[pagebackref,breaklinks,colorlinks,allcolors=cvprblue]{hyperref}
\usepackage{amssymb}

\usepackage{pifont}% http://ctan.org/pkg/pifont
\newcommand{\emoji}{\raisebox{-0.37\height}{\includegraphics[height=3.5\fontcharht\font`\B]{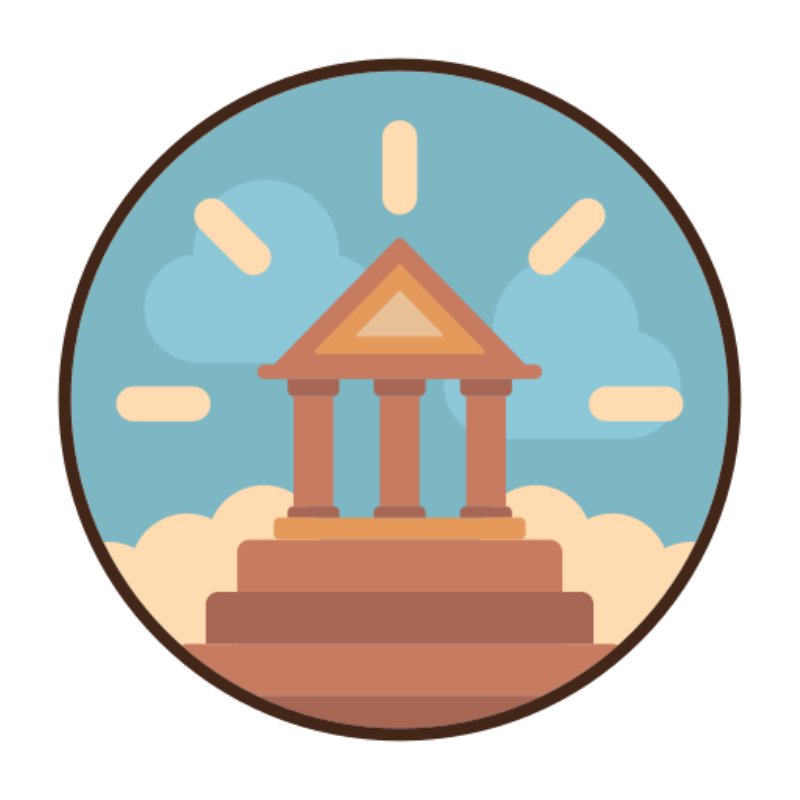}}}
\usepackage{url}
\usepackage{multirow} 
\usepackage{float}
\usepackage{color}
\usepackage{enumitem}
\usepackage{paralist}
\usepackage{booktabs}
\usepackage[T1]{fontenc}
\definecolor{blue_new}{rgb}{0.122, 0.318, 1}
\definecolor{deeppink}{rgb}{1.0, 0.08, 0.58}
\usepackage{array}
\hypersetup{
    colorlinks,
    linkcolor={red},
    citecolor={blue_new},
    urlcolor={deeppink}
}
\title{\centering \emoji{}\hspace{1.5mm}Olympus: A Universal Task Router for Computer Vision Tasks\vspace{-8mm}}

\author{
  \textbf{Yuanze Lin}$^\clubsuit$ \quad\quad \textbf{Yunsheng Li}$^\spadesuit$ \quad\quad \textbf{Dongdong Chen}$^\spadesuit$ \quad\\ \hspace{4mm} \textbf{Weijian Xu}$^\spadesuit$ \quad\quad \textbf{Ronald Clark}$^\clubsuit$ \quad\quad \textbf{Philip H. S. Torr}$^\clubsuit$ \quad\quad 
  \\
  $^\clubsuit$ University of Oxford \quad\quad $^\spadesuit$Microsoft\\
\url{http://yuanze-lin.me/Olympus_page/}
}

\begin{document}
\twocolumn[{%
\renewcommand\twocolumn[1][]{#1}%
\maketitle
\vspace{-3.2em} % Adjust space between title and figure if needed

\begin{center}
    \begin{minipage}{1.0\textwidth}
        \centering
        \includegraphics[width=\linewidth,keepaspectratio]{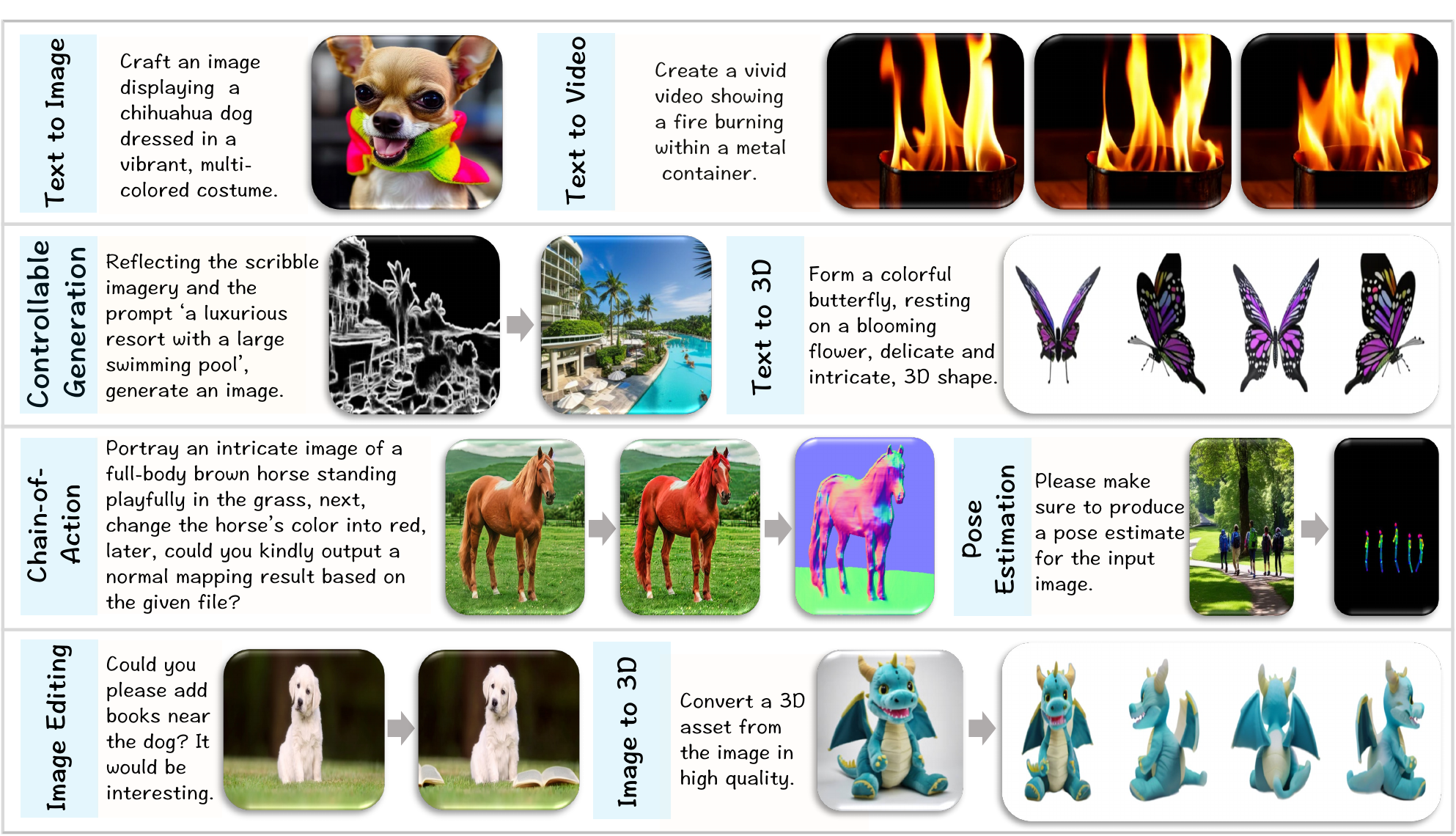}
        \vspace{-7mm} % Increase negative value to move caption upward
        \captionof{figure}{\small{\textcolor{black}{With its versatile capabilities, \textit{Olympus} addresses a broad spectrum of vision tasks across images, videos, and even 3D content, in fact, it can cover over 20 different tasks.}}}
        \label{figure1} % Place label immediately after caption
    \end{minipage}
\end{center}
\vspace{1em} % Adjust space after figure if needed
}]

\begin{abstract}
We introduce \textit{Olympus}, a new approach that transforms Multimodal Large Language Models (MLLMs) into a unified framework capable of handling a wide array of computer vision tasks. Utilizing a controller MLLM, \textit{Olympus} delegates over 20 specialized tasks across images, videos, and 3D objects to dedicated modules. This instruction-based routing enables complex workflows through chained actions without the need for training heavy generative models. \textit{Olympus} easily integrates with existing MLLMs, expanding their capabilities with comparable performance. Experimental results demonstrate that \textit{Olympus} achieves an average routing accuracy of 94.75\% across 20 tasks and precision of 91.82\% in chained action scenarios, showcasing its effectiveness as a universal task router that can solve a diverse range of computer vision tasks.
\end{abstract}

\section{Introduction}
\label{sec:intro}

\textit{``If I have seen further it is by standing on the shoulders of Giants."} -- Isaac Newton
\vspace{2mm}

Multimodal Large Language Models (MLLMs) have made significant strides in advancing understanding, generation and reasoning across diverse domains. For example, in understanding, MLLMs like LLaVA~\citep{liu2024visual} excel in visual question-answering (VQA)~\citep{antol2015vqa}, effectively integrating visual and textual data. In generation, diffusion models ~\citep{nichol2021improved,ho2020denoising,sdxl} have achieved exceptional results in text-to-image, text-to-video and text-to-3D generation~\citep{saharia2022photorealistic,singer2022make,sdxl,sd3,ho2022video,wu2023tune,chen2024text,lin2024dreampolisher,lin2024text}. 

Building on these advancements, recent studies~\citep{wu2023next, seed-x, x-vila, CoDI, dreamllm, lin2023smaug, xie2024show} have aimed to develop unified architectures capable of performing tasks across various domains. Notably, Emu3~\citep{wang2024emu3} and Omni-Gen~\cite{xiao2024omnigen} introduce all-in-one models designed to handle both generation and understanding tasks.  However, the integration of distinct domains within a single model continues to present significant challenges. In particular, variability within domains often leads to compromised performance on individual tasks due to conflicts between task objectives, such as those between text and image generation tasks~\cite{zhou2024transfusion}. These conflicts hinder the models' effectiveness and limit their utility in real-world applications. 

Another key limitation of all-in-one models lies in their constrained ability to handle a broad spectrum of vision-language tasks across different domains due to differing input and output formats. This restriction presents a substantial bottleneck to scalability, particularly as the range of tasks continues to grow across images, videos, and the emerging 3D domain. Furthermore, extending these models to accommodate new tasks is inherently challenging. Training such comprehensive models with increasing model sizes demands substantial computational resources, and highly complex training methodologies. For instance, Omni-Gen~\citep{xiao2024omnigen} necessitates 104$\times$A800 GPUs and five distinct training stages. These issues underscore the pressing need for modular or task-adaptive frameworks to enhance scalability and efficiency in addressing the increasingly diverse demands of vision tasks. Additionally, all-in-one models often struggle to integrate meticulously designed, task-specific components effectively, reducing their overall efficiency and performance in specialized applications.

This prompts us to explore another alternative approach for seamlessly unifying vision tasks within a single framework. Inspired by HuggingGPT~\citep{shen2024hugginggpt} and the advanced contextual understanding of MLLMs, we propose leveraging MLLMs to handle vision-language comprehension \textit{internally} while delegating other tasks \textit{externally}. Although technically straightforward, it represents a foundational and significant step toward advancing unified frameworks for computer vision tasks. Specifically, the MLLM can function as a task router, coordinating with specialized external models to address various tasks and overcome individual model limitations. However, it still faces significant challenges, primarily due to the variability in user prompts across a wide range of tasks and the lack of comprehensive, task-specific instruction datasets essential for effective training and evaluation.

% Our approach
In this paper, we introduce \textit{Olympus}, a unified framework that leverages MLLMs to handle a diverse array of computer vision tasks. Our \textit{Olympus} differs from existing methods~\citep{team2024chameleon, wang2024emu3, xie2024show, zhou2024transfusion} which focus on presenting all-in-one models to solve diverse tasks. To accomplish this, we collected \textbf{446.3K} high-quality training instructions and \textbf{49.6K} evaluation instructions from GPT-4o~\citep{hurst2024gpt} named as \textit{OlympusInstruct} and \textit{OlympusBench} respectively, spanning \textbf{20} different vision tasks. Furthermore, we designed specific routing tokens tailored to delegate individual tasks. Finally, leveraging these routing tokens and \textit{OlympusInstruct}, our model can even perform a chain of tasks within a single user instruction if needed.

\begin{figure*}
    \centering
    \includegraphics[width=1.01\linewidth]{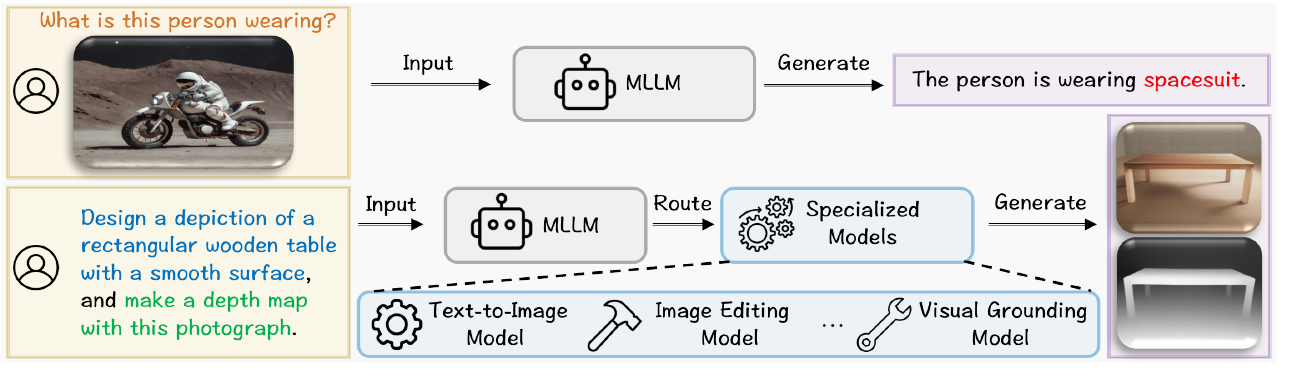}
    \vspace{-6mm}
    \caption{\small{\textcolor{black}{Given the user prompts, a trainable MLLM can perform routing across a wide range of specified models. In this concept, MLLMs can solve multimodal understanding tasks (e.g., VQA) with its inherited capacity, while MLLMs can allocate appropriate specialized models to address multimodal generative and classic vision tasks (e.g., image generation and depth estimation), then aggregate the results and deliver a response to the user.}}}
    \label{fig:figure2}
\end{figure*}

In our experiments, \textit{Olympus} achieves comparable performance to the leading MLLMs on standard multimodal benchmarks~\citep{llava}. Additionally, it supports over \textbf{20} distinct tasks across the domains of image, video, and 3D, as shown in Figure~\ref{figure1}. We further investigate the effectiveness of decomposing user instructions to interface with suitable external models, \textit{Olympus} achieves an impressive average routing accuracy of \textbf{94.75\%} across \textbf{20} individual tasks. In chain-of-action scenarios, which involves performing multiple tasks to complete an instruction, our model attains \textbf{91.82\%} precision. These results highlight the potential of \textit{Olympus}. In summary, our contributions can be included as:
\begin{compactitem}
    \item We introduce \textit{Olympus}, an innovative framework that leverages Multimodal Large Language Models (MLLMs) to perform contextual understanding tasks through their inherent capabilities, while addressing other tasks via allocating external models.
    
    \item We develop task-specific routing tokens and enhance MLLMs with chain-of-action capabilities. Our model achieves comparable performances with leading MLLMs on multimodal benchmarks, \textit{Olympus} achieves \textbf{94.75\%} routing accuracy in single-task scenarios, \textbf{91.82\%} precision in chain-of-action settings, and solves up to \textbf{5} tasks within a single instruction.
    
    \item We have curated high-quality instruction datasets named \textit{OlympusInstruct} and \textit{OlympusBench} across \textbf{20} computer vision tasks, comprising \textbf{446.3K} and \textbf{49.6K} samples for training and evaluation respectively. These datasets provide a solid foundation for further exploration and advancement in this domain.
\end{compactitem}

\section{Related Work}
\label{sec:related_work}
\subsection{Vision-Language Understanding}
Recent advancements in large language models (LLMs)~\citep{llama,gpt3} have catalyzed the development of multimodal large language models (MLLMs)~\citep{blip2,instructblip,alayrac2022flamingo,llava,llava1.5,qwen_vl,Kosmos-2,anil2023gemini,PALM,mckinzie2024mm1,tong2024cambrian,lin2024rethinking,li2024mini,li2024llava,liu2024llava,shi2024eagle,xue2024xgen}. Pioneering multimodal large language models (MLLMs), such as MiniGPT-4~\citep{minigpt4}, have demonstrated impressive capabilities in processing and integrating multiple modalities. Models like Kosmos-2~\citep{Kosmos-2}, LLaVA~\citep{llava} and LLaVA-OneVision~\citep{li2024llava} have further enhanced the visual cognitive abilities of MLLMs. Additionally, approaches including LLaVA-Phi~\citep{zhu2024llava}, MobileVLM~\citep{chu2023mobilevlm}, and Mipha~\citep{zhu2024mipha} focus on refining training methodologies and architectural frameworks to develop more efficient and lightweight MLLMs. Although these models excel in visual perception and multimodal understanding, they are predominantly limited to generating text-based outputs, which restricts their effectiveness across a broader range of vision tasks involving images, videos, and 3D content generation. In this work, we adopt a multimodal model structure following Mipha~\cite{zhu2024mipha}.

\begin{figure*}
    \centering
    \includegraphics[width=1.015\linewidth]{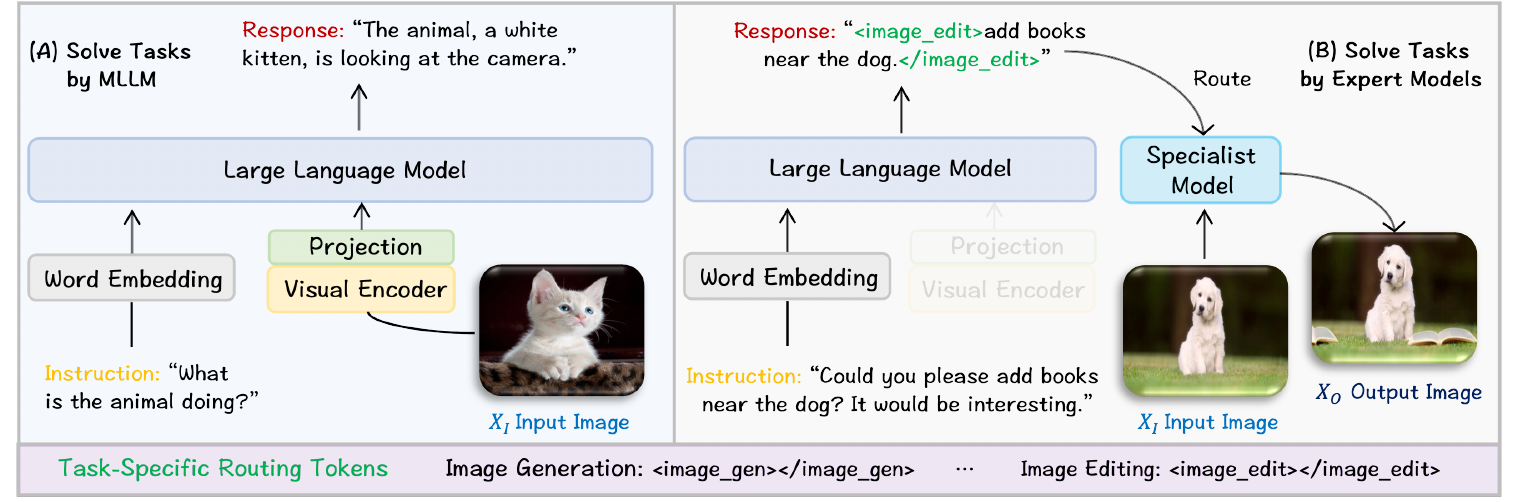}
    \vspace{-6mm}
    \caption{\small{\textcolor{black}{The framework of \textit{Olympus}. It can solve those tasks like VQA through the inherited capacities of MLLM directly. For other tasks, e.g., image editing, \textit{Olympus} can generate the response, which consists of task-specific routing tokens and refined prompts, they are then used to schedule specialist models for addressing diverse user requests.}}}
    \vspace{-1mm}
    \label{fig:figure3}
\end{figure*}

\subsection{Unified Vision-Language Foundation Model}
Extensive research~\citep{seed-x,seed-story,wu2023next,sun2023emu,wang2024emu3,xie2024show,CoDI,dreamllm,jointly,zhou2024transfusion,yu2023scaling,zhan2024anygpt} has focused on developing unified multimodal language models proficient in both understanding and generating content. Approaches such as~\citep{vlgpt,seed-x} integrate continuous embeddings with textual tokens within autoregressive frameworks for image generation. Emu2~\citep{sun2023emu} combines CLIP ViT~\citep{vit} image embeddings with text tokens for autoregressive modeling, while Chameleon~\citep{team2024chameleon} employs a transformer across diverse modalities with autogressive modeling. Show-o~\citep{xie2024show} and TransFusion~\citep{zhou2024transfusion} incorporate autoregressive and diffusion modeling within a single transformer. Omni-Gen~\citep{xiao2024omnigen} utilizes a VAE~\citep{vae} encoder-decoder alongside a transformer to process free-form prompts. Recently, Emu3~\cite{wang2024emu3} trained a unified transformer with next-token prediction across video, image, and text datasets, achieving superior performance on multimodal benchmarks and generation tasks. However, current unified multimodal foundation models predominantly support a narrow range of generative tasks, such as image and video creation or editing, and face significant scalability challenges for broader AI applications. Additionally, their training requires substantial computational resources. To overcome these limitations, we further enhance MLLMs by enabling the seamless integration of domain-specialized models tailored for diverse applications.

\subsection{LLM-Based Tools}
Large language models (LLMs)~\citep{llama}, trained on extensive datasets, demonstrate exceptional proficiency in zero-shot and few-shot settings, as well as in complex tasks such as mathematical problem-solving and commonsense reasoning. To extend their capabilities beyond text generation, recent research~\citep{schick2024toolformer,shen2024hugginggpt,suris2023vipergpt,wu2023visual,lin2022revive,liang2024taskmatrix,qin2023tool,shen2024hugginggpt} has focused on integrating external tools and models into LLM architectures. Toolformer~\citep{schick2024toolformer} pioneered this approach by embedding API calls within textual sequences, thereby enabling LLMs to utilize external tools effectively. Building upon this foundation, subsequent studies have incorporated visual modalities: Visual ChatGPT~\citep{wu2023visual} integrates LLMs with visual models such as BLIP~\citep{blip2}, Visual Programming~\citep{gupta2023visual} and ViperGPT~\citep{suris2023vipergpt} translate visual queries into executable Python code, facilitating the processing of visual data by LLMs. Additionally, HuggingGPT~\citep{shen2024hugginggpt} enhances large language models (LLMs) by utilizing them as controllers that direct user requests to specialized expert models, thereby integrating language comprehension with domain-specific expertise.

Although HuggingGPT assigns specific AI models to perform various tasks, it primarily relies on prompt engineering to leverage ChatGPT as an interface for connecting diverse external models without training. In contrast, our approach involves training multimodal large language models (MLLMs) from the ground up, enabling them to internally handle vision-language understanding tasks while designate specialized models to address a wide range of AI tasks.

\section{\textit{Olympus}}
\label{sec:method}
\textit{Olympus} leverages MLLMs as the foundation for various computer vision tasks. For vision-language tasks like visual question answering (VQA), MLLMs utilize their inherent capabilities. For other vision tasks, such as generative tasks (e.g., image, video, and 3D generation) and classic vision tasks (e.g., image super-resolution and depth estimation), MLLMs act as intermediaries, routing user instructions to specialized models. As shown in Figure~\ref{fig:figure2}, upon receiving a request, the MLLM can autonomously orchestrate the workflow, coordinating expert models to achieve the objective. The following subsections outline the details of \textit{Olympus}.

\subsection{Instruction Dataset Collection}
\label{data}
\begin{figure*}
    \centering
    \includegraphics[width=1\linewidth]{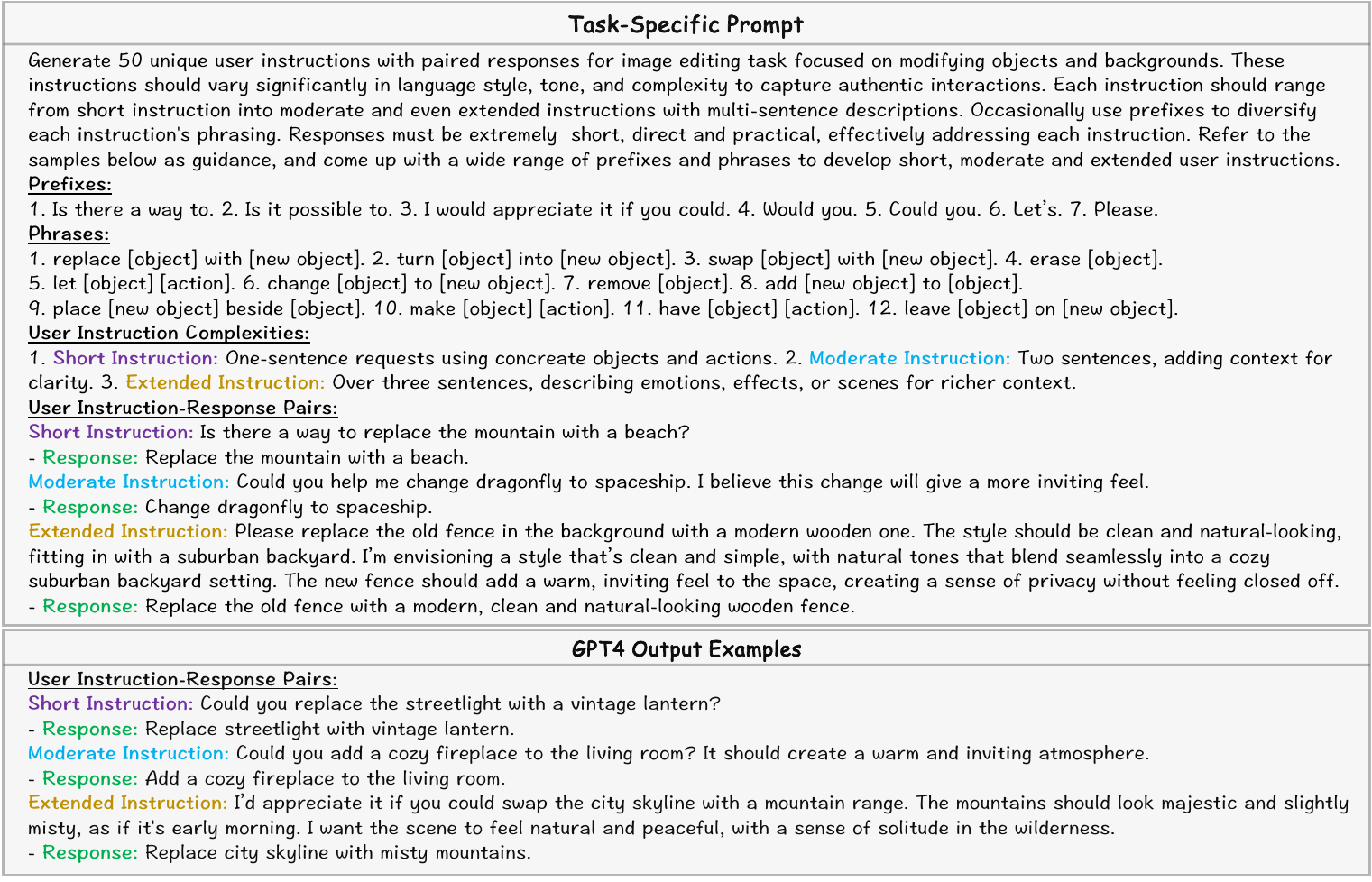}
    \vspace{-7mm}
    \caption{\small{\textcolor{black}{One example to illustrate the prompt we use to generate the instruction-response pairs for image editing by GPT-4o.}}}
    \label{fig:figure4}
    \vspace{-2mm}
\end{figure*}

In order to accurately assign user instructions to the appropriate model, we constructed a high-quality and diverse dataset of user instruction–response pairs using GPT-4o. This dataset comprises \textbf{446.3K} training samples, designated as \textit{OlympusInstruct}, and \textbf{49.6K} evaluation samples, designated as \textit{OlympusBench}, encompassing \textbf{20} distinct tasks. For each task, a specialized prompt was developed to align with the specific context of the task. This involved crafting detailed directives that enable GPT-4o to generate coherent and contextually relevant user requests and responses. An example of the image editing prompt used by GPT-4o to collect user instruction–response pairs is provided in Figure~\ref{fig:figure4}.

To ensure diversity in user instructions, we incorporated various prefixes and phrases that introduce different language styles, tones, and structures. Additionally, we categorized instruction complexities into three levels: short, moderate, and extended. This stratification allows GPT-4o to produce instructions that vary in length and complexity. Furthermore, we prompted GPT-4o to generate responses that are both practical and direct, thereby enhancing the applicability and clarity of the interactions. We also performed thorough data cleaning by removing duplicate entries and utilizing GPT-4o to remove entries that were contextually or grammatically inappropriate. This purification process ensures the integrity and quality of the dataset. Figure~\ref{fig:figure4} presents examples for image editing task, demonstrating how the task-specific prompts enable GPT-4o to generate instructions with diverse levels of complexity and varied language styles.

\begin{figure*}[!t]
    \centering
    \includegraphics[width=1.24\linewidth]{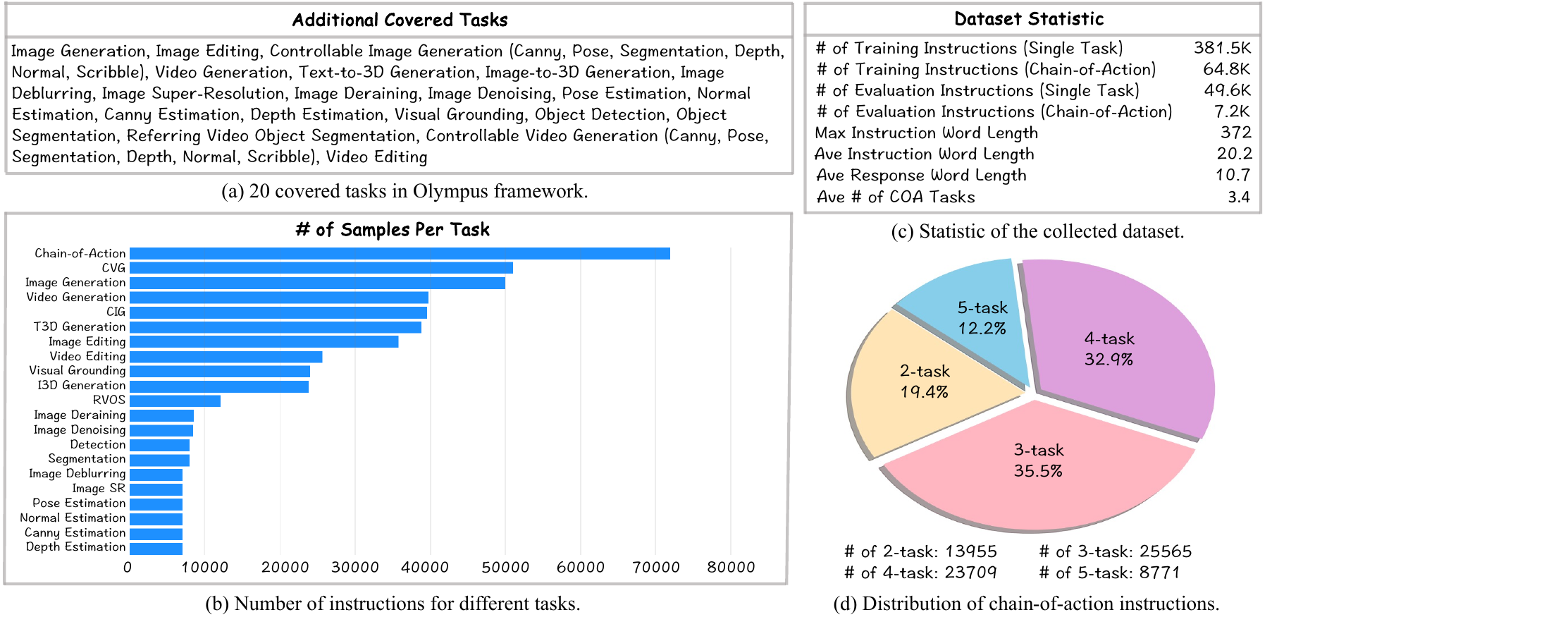}
    \vspace{-8mm}
    \caption{\small{\textcolor{black}{The statistic of the collected dataset. Note that CVG and CIG denote controllable video generation and controllable image generation, RVOS represents referring video object segmentation and image SR means image super-resolution in figure (b).}}}
    \label{fig:figure5}
    \vspace{-2mm}
\end{figure*}

Figure~\ref{fig:figure5} illustrates the statistical characteristics of our training and evaluation datasets. Specifically, the training set comprises \textbf{381.5K} single-task instruction–response pairs and \textbf{64.8K} chain-of-action instruction–response pairs. Similarly, the evaluation set consists of \textbf{49.6K} single-task pairs and \textbf{7.2K} chain-of-action pairs. The maximum word length across all instructions is \textbf{372} words, with an average instruction length of \textbf{20.2} words. Responses have an average length of \textbf{10.7} words. The \textbf{20} covered tasks are categorized into three groups: \textbf{(1) Image domains:} image generation/editing, deblurring, deraining, super-resolution, denoising, pose/normal/canny/depth estimation, controllable image generation across six conditions (canny, pose, segmentation, depth, normal, scribble), object detection/segmentation, and visual grounding; \textbf{(2) Video domains:} video generation/editing, controllable video generation across the same six conditions, and referring video object segmentation; and \textbf{(3) 3D domains:} image-to-3D and text-to-3D generation.

\subsection{Task-Specific Routing Tokens}
\label{tokens}
As shown in Figure~\ref{fig:figure3}, \textit{Olympus} directs user requests to dedicated models using task-specific routing tokens (e.g., image editing). To facilitate MLLMs in predicting appropriate models aligned with users' goals, we design a set of routing tokens specific to individual tasks. For instance, in the domains of image and video generation, we use the routing tokens \verb|<image_gen>|$\cdots$\verb|</image_gen>| and \verb|<video_gen>|$\cdots$\verb|</video_gen>|, respectively. Given a user instruction such as \texttt{"Please craft an image displaying a chihuahua dog dressed in a vibrant, multicolored costume."}, the corresponding response can be \verb|<image_gen>|\texttt{a chihuahua dog dressed in a vibrant, multicolored costume.}\verb|</image_gen>|, which is designed to effectively address the user's request. Thus, user instructions and the responses form input-answer pairs for training. Detailed information on designed routing tokens and corresponding specialist models are explained in the \textbf{Appendix}.

\textbf{Chain-of-Action. } By introducing domain-specific routing tokens, \textit{Olympus} enables chain-of-action capabilities, allowing to handle multiple tasks within a single instruction. For instance, consider a user prompt combining pose-based image generation and image editing: \texttt{"In homage to the pose imagery and the prompt 'a majestic castle', generate an image. In the following step, please refine the image by adding green trees."} The predicted response using the routing tokens is: \verb|<pose_to_image>|\texttt{a majestic castle}\verb|</pose_to_image>|\verb|<image_edit>|\texttt{adding green trees}\verb|</image_edit>|. Therefore, \textit{Olympus} can sequentially route user instructions to the appropriate modules for pose-conditioned image generation and image editing, in alignment with the task-oriented routing tokens. Moreover, \textit{Olympus} supports up to five consecutive tasks within a single prompt and is capable of scaling to accommodate an even larger number of tasks, thereby demonstrating its flexibility and scalability.

\begin{table*}[t]
  \centering
  \resizebox{1.0\linewidth}{!}{
      \begin{tabular}{l|l|l|ccccccccccc}
           \textbf{Method} & \textbf{LM}  & \textbf{Res.} & $\textbf{\text{VQAv2}}$ &\textbf{\text{GQA}} & \textbf{\text{VisWiz}} & \textbf{\text{SQA$^{\text{I}}$}} & \textbf{$\text{VQA}^{\text{T}}$} & \textbf{\text{MME-P}} & \textbf{\text{MME-C}} & \textbf{\text{MMB}} & \textbf{\text{MM-Vet}} & \textbf{\text{POPE}} & \textbf{\text{MMMU}} \\
        \midrule
        Shikra~\cite{chen2023shikra} & V-13B & 224 &  77.4 & -& - & - & - & - & - & 58.8 & - & -& -\\
        IDEFICS-9B~\cite{laurenccon2024obelics}& L-7B & 224 & 50.9 & 38.4 & 35.5 & - & 25.9 & - & - & 48.2 & - & -  & - \\
        IDEFICS-80B~\cite{laurenccon2024obelics}& L-65B & 224 &  60.0 & 45.2 & 36.0 & - & 30.9 & - & - & 54.5 & - & - & - \\
        Qwen-VL-Chat~\cite{qwen_vl} & Q-7B & 448   & 78.2 & 57.5 &38.9 & 68.2 & 61.5 & 1487.5 & 360.7 & 60.6 & - & - & 32.9\\
        mPLUG-Owl2~\cite{ye2023mplug} & L-7B & 448 &  79.4 & 56.1 &54.5 & 68.7 & 58.2 & 1450.2 & 313.2 & 64.5 & 36.2 & 85.8 & 32.1\\
        LLaVA-1.5~\cite{llava1.5} & V-7B & 336 &  78.5 & 62.0 & 50.0 & 66.8 & 58.2 & 1510.7 & 316.1 & 64.3 & 30.5 & 85.9 & 32.0\\
        MobileVLM-3B ~\cite{chu2023mobilevlm}& M-2.7B & 336 & - & 59.0 & - & 61.2 & 47.5 & 1288.9 & - & 59.6 & - & 84.9 & - \\
        MobileVLM-v2-3B~\cite{chu2024mobilevlm} & M-2.7B & 336 & -  & 61.1 & - & 70.0 & 57.5 & 1440.5 & - & 63.2 & - & 84.7 & -\\
        LLaVA-Phi~\cite{zhu2024llava} & P-2.7B & 336  & 71.4 & -  & 35.9 & 68.4 & 48.6 & 1335.1 & - & 59.8 & 28.9 & 85.0 & -\\
        Imp-v1~\cite{sung2023empirical} & P-2.7B & 384 & 79.5 & 58.6  & - & 70.0 & 59.4 & 1434.0 & - & 66.5 & 33.1 & 88.0 & -\\
        MoE-LLaVA-3.6B~\cite{lin2024moe} & P-2.7B & 384 & 79.9 & 62.6 & 43.7 & 70.3 & 57.0 & 1431.3 & - & 68.0 & 35.9 & 85.7 & -\\
        TinyLLaVA~\cite{zhou2024tinyllava} & P-2.7B & 384 & 79.9 & 62.0 & - & 69.1 & 59.1 & 1464.9 & - & 66.9 & 32.0 & 86.4 & -\\
        Bunny-3B~\cite{he2024efficient} & P-2.7B & 384 & 79.8 & 62.5 & - & 70.9 & - & 1488.8 & 289.3 & 68.6 & - & 86.8 & 33.0\\
        Mipha-3B~\cite{zhu2024mipha} & P-2.7B  & 384 &  81.3 & 63.9 & 45.7 & 70.9 & 56.6 & 1488.9 & 295.0 & 69.7 & 32.1 & 86.7 & 32.5\\
       \midrule
        \rowcolor{LightCyan} \textbf{\textit{Olympus} (Ours)} & P-2.7B  & 384 & 80.5 & 63.9 & 48.2 & 70.7 & 53.4 & 1520.7 & 283.2 & 71.2 & 33.8 & 86.6 & 32.8 \\
      \end{tabular}
    }
\vspace{-2mm}
\caption{Multimodal evaluation across $11$ benchmarks: $\text{VQA}{\text{v2}}$~\cite{goyal2017making}, \text{GQA}~\cite{hudson2019gqa}, \text{VisWiz}~\cite{gurari2018vizwiz}, $\text{SQA}^{\text{I}}$: ScienceQA-IMG~\cite{lu2022learn}, $\text{VQA}^{\text{T}}$: TextVQA~\cite{singh2019towards}, \text{MME-P}: MME Perception~\cite{mme}, \text{MME-C}: MME Cognition~\cite{mme}, \text{MMB}: MMBench~\cite{liu2023mmbench}, \text{MM-Vet}~\cite{yu2023mm}, \text{POPE}~\cite{li2023evaluating} and \text{MMMU}~\cite{yue2024mmmu}. “V”, “L”, “Q”,  “M” and “P” represent Vicuna~\cite{vicuna}, LLaMA~\cite{llama}, Qwen~\cite{qwen_vl}, MobileLLaMA~\cite{chu2023mobilevlm} and Phi-2~\cite{microsoft2024phi2}. {Res.} refers to the image resolution used by the visual backbone.} 
\label{table1}
\vspace{-2mm}
\end{table*}

\subsection{Training}
\label{training}
Since the goal is to generate task-specific response together with its routing tokens conditioning on the user instructions, we can train MLLMs with next-token prediction paradigm using the cross-entropy loss:

\begin{equation}
P(Y_{\mathrm{a}} | \mathcal{F}_{\mathrm{v}}, \mathcal{F}_{\mathrm{t}}) = \prod_{i=1}^{L} P_{\boldsymbol{\theta}} (y_{i} | \mathcal{F}_{\mathrm{v}}, \mathcal{F}_{\mathrm{t}}, Y_{\mathrm{a},<i}).
\end{equation}
Here, $L$ represents the sequence length of the response $Y_{\mathrm{a}}$, $\mathcal{F}_{\mathrm{v}}$ denotes the visual embedding which is adopted for those multimodal instructions,  and $\boldsymbol{\theta}$ means the trainable parameters of MLLMs. The notation $Y_{\mathrm{a},<i}$ denotes all tokens preceding the current token $y_{i}$, and $\mathcal{F}_{\mathrm{t}}$ represents the input instruction embeddings.

\subsection{Inference}
As displayed in Figure \ref{fig:figure3}, upon receiving a prompt, \textit{Olympus} generates a response with task-customized routing tokens. These tokens invoke the appropriate AI models to handle various tasks, and their predictions are aggregated into a final response. For tasks solvable by MLLMs alone, responses are generated directly, bypassing routing tokens.

\textit{Olympus} introduces a training-based paradigm where MLLMs act as controllers, addressing diverse tasks using both internal capabilities and external expert models. This general framework seamlessly integrates MLLMs with domain-specific models to tackle universal tasks.

\section{Experiment}
\label{sec:experiment}

\begin{figure*}[!t]
    \centering
    \includegraphics[width=\textwidth]{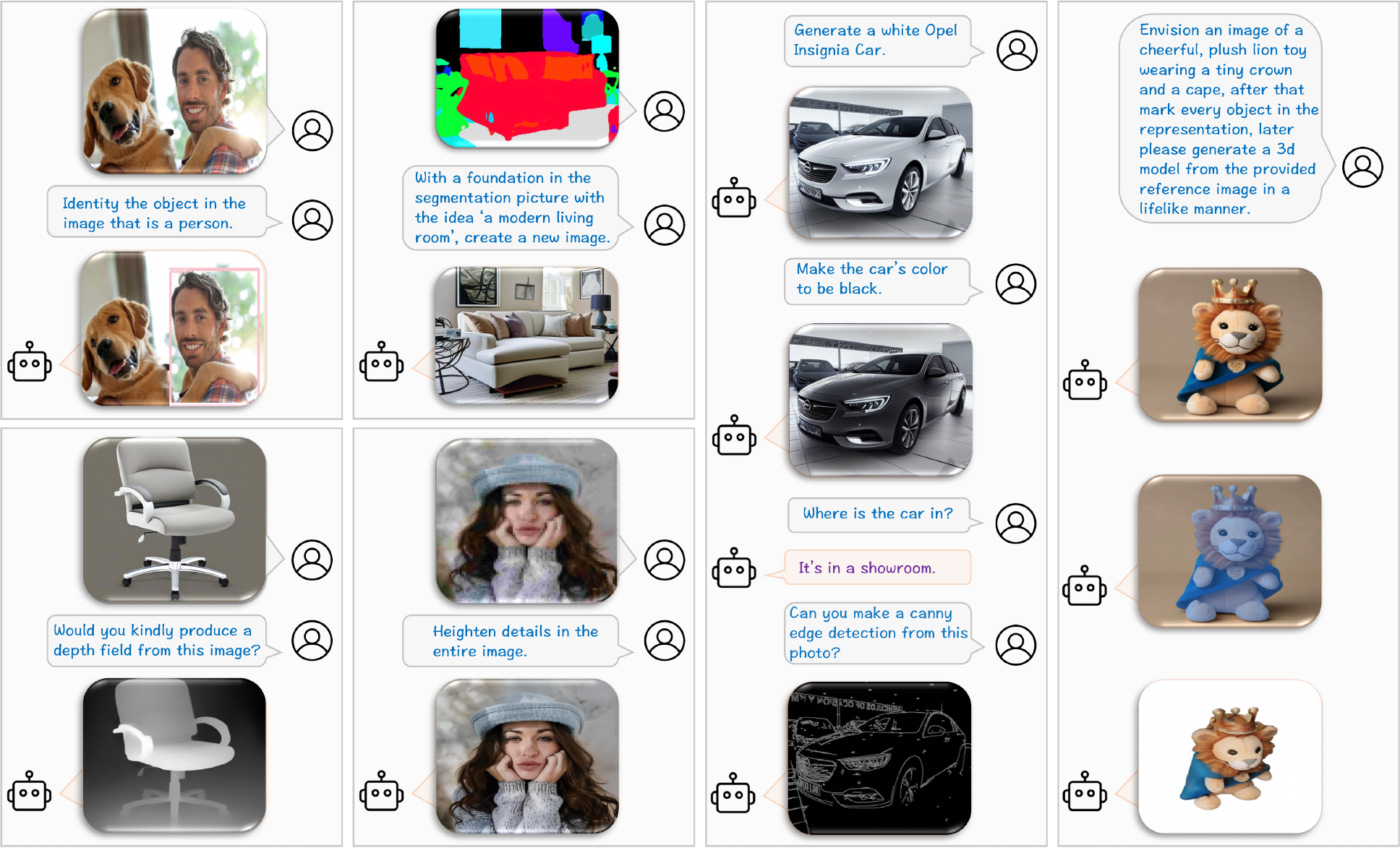}
    \vspace{-6mm}
    \caption{\small Diverse applications of \textit{Olympus}. The \textbf{\textcolor{orange}{1st}} and \textbf{\textcolor{blue}{2nd}} columns show the scenarios for single task, the \textbf{\textcolor{violet}{3rd}} column displays the results under multi-turn conversations, while \textbf{\textcolor{red}{the last (4th)}} column shows the chain-of-action capacity of \textit{Olympus}.}
    \label{fig:figure6}
    \vspace{-4mm}
\end{figure*}

\subsection{Experimental Setup}
\textbf{MLLM Model}. Our model follows the setting of Mipha~\citep{zhu2024mipha} with its vision and language encoders, i.e., SigLIP-384~\cite{zhai2023sigmoid} and Phi-2~\citep{microsoft2024phi2}. For the multimodal projector, same as LLaVA~\citep{llava} and Mipha~\citep{zhu2024mipha}, we adopt a two-layer MLP.

\textbf{Training Setting.} We initialize the weights from Mipha-3B~\citep{zhu2024mipha} and jointly fine-tune the model on both the LLaVA-Mix665K dataset~\citep{llava} and \textit{OlympusInstruct} for 2 epochs, using a learning rate of 5e-5 and a batch size of 256 on 64 V100 32GB GPUs. The whole training process takes approximately 24.8 hours. All model components, including the vision encoder, language encoder, and MLP, are fully fine-tuned during the training process.

\textbf{Evaluation Details.} 
We compare our method with a bunch of state-of-the-art multimodal large language models (MLLMs) across 11 popular benchmarks, as shown in Table~\ref{table1}. These benchmarks include VQA-v2~\citep{goyal2017making}, GQA~\citep{hudson2019gqa}, ScienceQA-IMG~\citep{lu2022learn}, MME perception and cognition~\citep{mme}, MMBench~\citep{liu2023mmbench}, MM-Vet~\citep{yu2023mm}, TextVQA~\citep{singh2019towards}, and POPE~\citep{li2023evaluating}, etc. For task routing performance, we evaluate accuracy, precision, recall, and F1 score, and for chain-of-action tasks, we also report edit distance~\citep{marzal1993computation}, following HuggingGPT~\citep{shen2024hugginggpt}.

\begin{table}[!t]
  \centering
  \footnotesize
  \resizebox{\linewidth}{!}{
      \begin{tabular}{l|cccc}
           \textbf{Method}  & \textbf{Acc} $\uparrow$ & \textbf{Pre} $\uparrow$ & \textbf{Recall} $\uparrow$ & \textbf{F1} $\uparrow$ \\
        \midrule
        HuggingGPT (GPT-4o mini) & 70.14 & 76.51 & 72.14 & 75.46 \\
        HuggingGPT (GPT-4o) & 81.35 & 85.54 & 81.55 & 83.56 \\
        \rowcolor{LightCyan} \textbf{Olympus (Ours)} & \textbf{94.75} & \textbf{95.80} & \textbf{94.75} & \textbf{95.77}
      \end{tabular}
  }
  \vspace{-2mm}
  \caption{Evaluation results on \textit{\textit{Olympus}Bench} under the single-task setting. Metrics include accuracy (\%), precision (\%), recall (\%), and F1-score (\%).}
  \label{table2}
  \vspace{-2mm}
\end{table}

\begin{table}[!t]
  \centering
  \footnotesize
  \resizebox{\linewidth}{!}{
      \begin{tabular}{l|cccc}
           \textbf{Method}  & \textbf{ED} $\downarrow$ & \textbf{Pre} $\uparrow$ & \textbf{Recall} $\uparrow$ & \textbf{F1} $\uparrow$ \\
        \midrule
        HuggingGPT (GPT-4o mini) & 0.45 & 65.14 & 48.51 & 53.14 \\
        HuggingGPT (GPT-4o) & 0.35 & 75.03 & 60.23 & 61.25 \\
        \rowcolor{LightCyan} \textbf{Olympus (Ours)} & \textbf{0.18} & \textbf{91.82} & \textbf{92.75} & \textbf{91.98} \\
      \end{tabular}
  }
  \vspace{-2mm}
  \caption{Evaluation results on \textit{\textit{Olympus}Bench} under the chain-of-action setting. ED represents edit distance.}
  \label{table3}
  \vspace{-5mm}
\end{table}

\subsection{Quantitative Evaluation}
\textbf{Task Routing Performance.} To demonstrate our routing effectiveness, we compare our results with HuggingGPT on \textit{OlympusBench} using GPT-4o mini and GPT-4o models for their strong predictive capabilities in Table~\ref{table2} and \ref{table3}. For a fair comparison, we included prompts covering all task types supported by \textit{Olympus} and excluded prompts for irrelevant tasks. In Table~\ref{table2}, under the single-task setting, our method achieves notable improvements of \textbf{13.4\%}, \textbf{10.26\%}, \textbf{13.2\%}, and \textbf{12.21\%} in accuracy, precision, recall, and F1 score, even against the strong GPT-4o model. In the chain-of-action setting, \textit{Olympus} demonstrates further gains of \textbf{0.17}, \textbf{16.79\%}, \textbf{32.52\%}, and \textbf{30.73\%} for edit distance, precision, recall, and F1 score, respectively.

\begin{table}[t]
  \centering
  \footnotesize
  \resizebox{0.75\linewidth}{!}{
      \begin{tabular}{l|c}
           \textbf{Method} & \textbf{Success Rate} $\uparrow$   \\
        \midrule
        HuggingGPT (GPT-4o mini) & 65.8 \\
        HuggingGPT (GPT-4o) & 75.2\\
        \rowcolor{LightCyan} \textbf{Olympus (Ours)} & \textbf{86.5} \\
      \end{tabular}
    }
\vspace{-2mm}
  \caption{Human Evaluation of different methods, we display the results of success rate (\%).}
\vspace{-5mm}
\label{table5}
\end{table}

\begin{table*}[!t]
  \centering
  \footnotesize
  \label{tbl:full_table}
  \resizebox{0.95\linewidth}{!}{
      \begin{tabular}{c|ccccccccccc}
        \textbf{\# of Tasks} & \textbf{VQAv2} & \textbf{GQA} & \textbf{VisWiz} & \textbf{SQA$^{\text{I}}$} & \textbf{VQA$^{\text{T}}$} & \textbf{MME-P} & \textbf{MME-C} & \textbf{MMB} & \textbf{MM-Vet} & \textbf{POPE} & \textbf{MMMU} \\
        \midrule
        0  &  81.0  &  64.0  & 46.2   & 70.8   &  55.3  & 1498.3   &  293.2  &  70.1  & 32.6  &  86.6  & 32.4   \\
        5  &  80.5  &  64.2  & 45.6   &  70.9  & 53.5   & 1468.3   &  310.4  &  70.5  &  34.9  & 86.5   &  32.5  \\
        10 &  80.4  & 64.1   &  46.1  &  71.2  &  53.0  &  1546.7   &  333.9  &  70.2  & 33.8   & 86.2   &  32.9  \\
        \rowcolor{LightCyan}\textbf{20} & 80.5 & 63.9 & 48.2 & 70.7 & 53.4 & 1520.7 & 283.2 & 71.2 & 33.8 & 86.6& 32.8 \\
      \end{tabular}
  }
  \vspace{-2mm}
    \caption{The ablation study of varying the number of tasks on multimodal benchmarks. ``0 task'' denotes fine-tuning the model only using LLaVA-Mix665K dataset~\citep{llava1.5}.}
  \label{table4}
\vspace{-5mm}
\end{table*}

Additionally, we collected 200 diverse human-generated instructions to evaluate \textit{Olympus}'s real-life performance against HuggingGPT, using success rate as the metric. The success rate reflects whether specialized models generate outputs that fully satisfy user requests, requiring both accurate task planning and effective task-tailored prompt generation. As shown in Table~\ref{table5}, \textit{Olympus} outperforms HuggingGPT with an \textbf{11.3\%} higher success rate using GPT-4o. These results highlight the significant potentials of \textit{Olympus} and the collected \textit{OlympusInstruct} dataset.

\textbf{Multimodal Understanding Performance.} Table~\ref{table1} compares our method with existing approaches on multimodal benchmarks. \textit{Olympus} achieves comparable performance to Mipha-3B on multiple benchmarks and even surpass it on 4 benchmarks such as VizWiz (\textbf{+2.5\%}), MME-P (\textbf{+31.8}), MMB (\textbf{+1.5\%}), and MM-Vet (\textbf{+1.7\%}). While it shows a slight drop on TextVQA, \textit{Olympus} uniquely supports model routing for 20 diverse vision tasks.

\begin{table}[!t]
  \centering
  \footnotesize
  \resizebox{0.85\linewidth}{!}{
      \begin{tabular}{c|ccccc}
           \textbf{\# of tasks} & \textbf{Acc} $\uparrow$ & \textbf{Pre} $\uparrow$   & \textbf{Recall} $\uparrow$ & \textbf{F1} $\uparrow$ \\
        \midrule
        5 & 96.38 & 96.36 & 96.45 & 97.61 \\
        10 & 96.15 & 95.85 & 96.23 & 97.07 \\
        15 & 95.84 & 95.78 & 95.84 & 96.79 \\
        \rowcolor{LightCyan} \textbf{20} & 94.75 & 95.80 & 94.75 & 95.77 \\
      \end{tabular}
    }
\vspace{-2mm}
\caption{Ablation study of different numbers of tasks adopted for training under single-task setting.}
\vspace{-5mm}
\label{table6}
\end{table}

\subsection{Ablation Study} 
\begin{figure}[!t]
    \centering
    \includegraphics[width=0.43\textwidth]{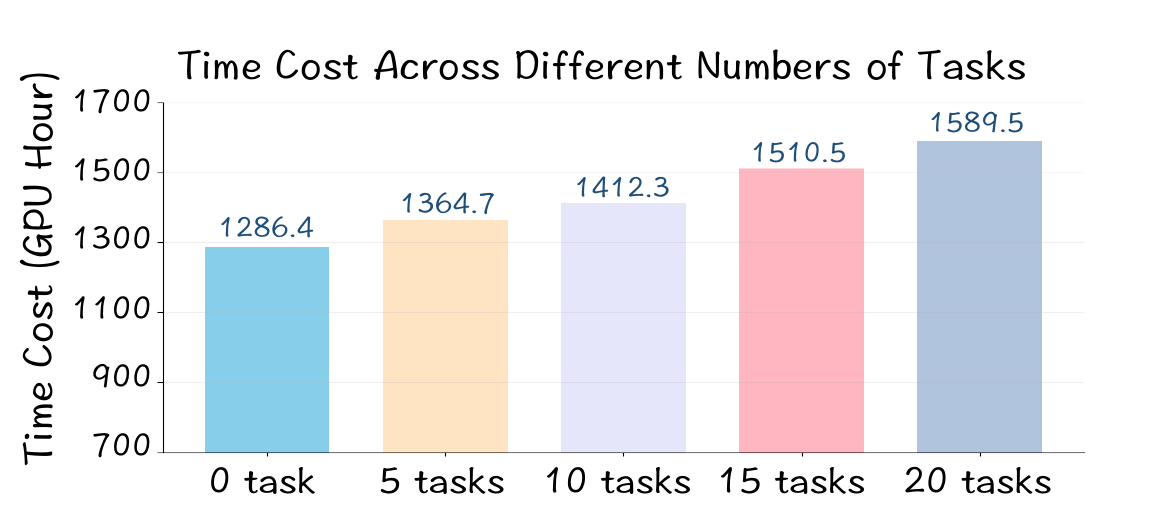}
    \vspace{-4mm}
    \caption{\small Training time cost for varying numbers of tasks.}
    \label{fig:figure7}
\vspace{-5mm}
\end{figure}

The ablation study exploring the impact of varying the number of training tasks is presented in Tables~\ref{table4}, \ref{table6}, \ref{table7}, and illustrated in Figure~\ref{fig:figure7}. Table~\ref{table4} demonstrates that the number of tasks has a limited influence on overall performance across multimodal benchmarks. Notably, the 10-task setting achieves the best results on MME-P~\cite{mme}, while the 20-task setting performs optimally on VizWiz~\cite{gurari2018vizwiz}. The 20-task configuration is selected for its robustness and generality across a diverse range of tasks. Tables~\ref{table6} and \ref{table7} illustrate a slight performance degradation in both single-task and chain-of-action settings as the number of tasks increases. This degradation is reasonably attributed to the increased prediction complexity associated with handling a larger number of tasks.

Figure~\ref{fig:figure7} highlights the training cost, which increases from 1286.4 GPU hours without utilizing \textit{OlympusInstruct} to 1589.5 GPU hours with it, representing a modest 23.6\% increase in time. This relatively low cost increase is attributed to the avoidance of training complex generative models. Further experimental details are provided in the \textbf{Appendix}.

\subsection{Visualization}
Figure~\ref{fig:figure6} illustrates the versatility of \textit{Olympus} across various tasks. The first two columns present single-turn examples, including visual grounding, depth estimation, controllable image generation, and image super-resolution. The third column illustrates \textit{Olympus}'s proficiency in executing a variety of tasks, such as image editing, visual question answering (VQA), and canny edge detection, within the context of multi-turn conversations. This is particularly noteworthy given that \textit{OlympusInstruct} does not include any multi-turn conversation data, underscoring our model's impressive capacity for generalization. The final column showcases its chain-of-action capability to conduct text-to-image generation, object segmentation and image-to-3D generation within one instruction. These examples clearly show that \textit{Olympus} can handle diverse prompts for multiple tasks and generate comprehensive responses for users.

\section{Limitation}
Since \textit{Olympus} is trained on the dataset collected through GPT-4o, it still has some limitations, e.g., the quality and diversity of the samples collected directly impact the performance of the generated responses. The inherent biases and inaccuracies in GPT-4o's responses propagate into MLLMs, potentially leading to suboptimal or biased outputs.

\begin{table}[!t]
  \centering
  \footnotesize
  \resizebox{0.85\linewidth}{!}{
      \begin{tabular}{c|ccccc}
           \textbf{\# of tasks} & \textbf{ED} $\downarrow$ & \textbf{Pre} $\uparrow$   & \textbf{Recall} $\uparrow$ & \textbf{F1} $\uparrow$ \\
        \midrule
        5 & 0.12 & 93.23 & 94.32 & 93.35\\
        10 & 0.14 & 92.23 & 93.45 & 92.28\\
        15 & 0.17 & 91.97 & 92.89 & 92.01\\
        \rowcolor{LightCyan}\textbf{20} & 0.18 & 91.82 & 92.75 & 91.98\\
      \end{tabular}
    }
\vspace{-2mm}
\caption{Ablation study of different numbers of tasks adopted for training under chain-of-action setting.}
\vspace{-5mm}
\label{table7}
\end{table}

\section{Conclusion}
We present \textit{Olympus}, a universal task router designed to address diverse computer vision tasks by integrating MLLM's internal abilities with task-specific routing to expert models. To achieve this, we introduce \textit{OlympusInstruct} and \textit{OlympusBench}, datasets collected from GPT-4o covering 20 distinct tasks. With the presented routing tokens, \textit{Olympus} can handle multiple tasks within a single prompt, highlighting its potential as a robust foundation for unifying a wide range of computer vision tasks.
{
    \small
    \bibliographystyle{ieeenat_fullname}
    \bibliography{main}
}
\clearpage
\appendix
\section{Appendix}
 \label{appendix}
In the supplementary materials, we provide the following sections:

\begin{compactitem}
\item Training details in Section \ref{implementation}.
\item Task-specific routing tokens in Section \ref{tokens2}.
\item Adopted specialist models in Section \ref{models}.
\item More dataset statistic in Section \ref{dataset}. 
\item Ablation study experiments in Section \ref{ablation}. 
\item More results in Section \ref{visualization}.
\end{compactitem}

\section{Training Details}
\label{implementation}
We train our models using 64 $\times$ V100 GPUs, each equipped with 32GB of memory. The Adam optimizer~\citep{kingma2014adam} is employed, combined with a cosine learning rate scheduler, aligning with the configuration utilized in LLaVA~\citep{llava}. For fine-tuning, we set a learning rate of 5e-5, which is optimized for stability and convergence, and adopt a batch size of 256 to accommodate the large-scale data and distributed training setup. The training process spans two epochs over the combined fine-tuning datasets of LLaVA-Instruct-158K~\citep{llava1.5} and \textit{OlympusInstruct}, ensuring that the models are effectively exposed to both general-purpose and task-specific instructions. 

Additional training configurations include a warmup ratio of 0.03, which helps in stabilizing the initial training phase, and gradient accumulation steps set to 4 to balance memory efficiency with gradient updates. To handle varying image sizes in the datasets, we employ a padding-based image aspect ratio strategy. Moreover, numerical precision is set to float16, enabling faster computation and reduced memory usage while maintaining sufficient numerical accuracy. These hyperparameters, summarized in Table~\ref{table:sup1}, were meticulously selected to optimize training performance and ensure scalability across diverse tasks.
\begin{table}[h]
    \centering
    \setlength{\tabcolsep}{0.15cm}
    \renewcommand{\arraystretch}{1.2}
    \resizebox{0.8\linewidth}{!}{
    \begin{tabular}{l c}
         \toprule
         Configuration                      & MLLMs Traning  \\
         \midrule
         Optimizer & Adam \\
         Learning rate                          & 5e-5 \\
         Learning rate schedule             & cosine \\
         Total training epochs               & 2 \\
         Weight Decay                       & 0 \\
         Warmup ratio                       & 0.03 \\
         Batch size                         & 256 \\
         Gradient Accumulation Steps & 4 \\
         Image Aspect Ratio & Pad \\
         Numerical precision                & $\mathtt{float16}$ \\
         \bottomrule
    \end{tabular}
  }
  \caption{Summary of training hyperparameters of \textit{Olympus}.}
  \label{table:sup1}
\end{table}

\section{Task-specific Routing Tokens}
\label{tokens2}
\begin{figure*}[!t]
    \centering
    \includegraphics[width=1\linewidth]{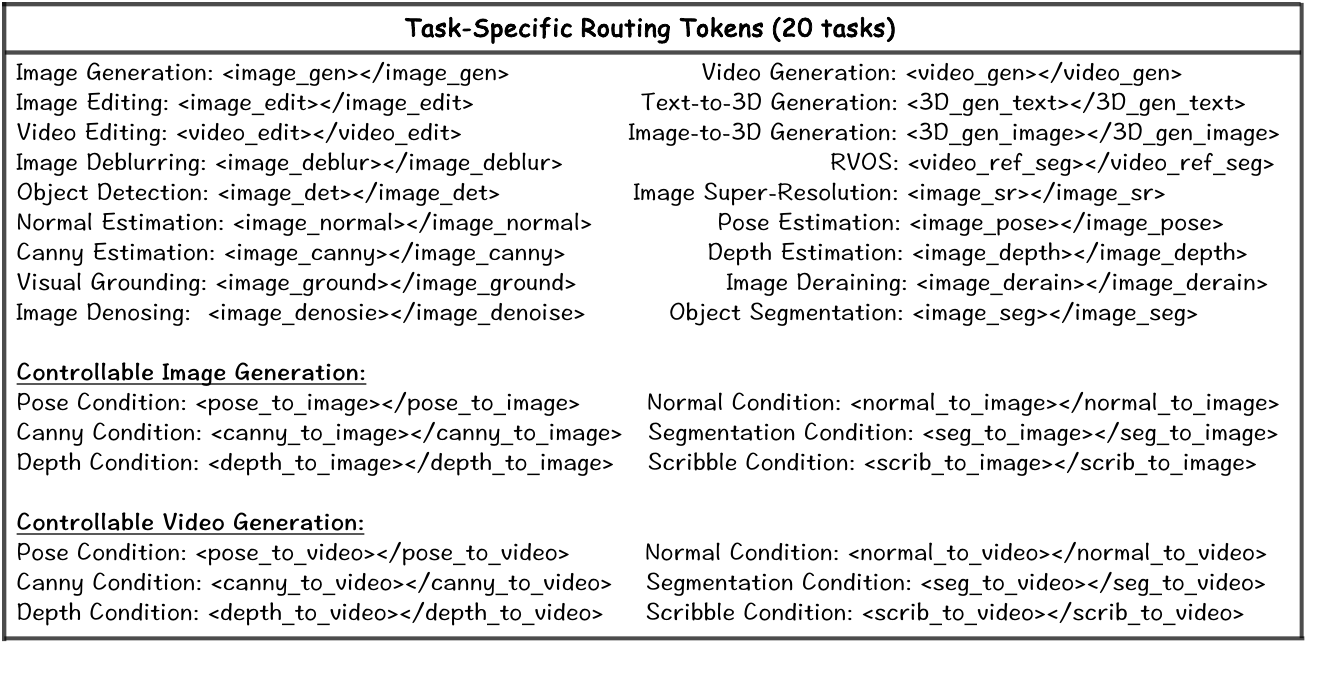}
    \vspace{-8mm}
    \caption{\small{\textcolor{black}{Task-specific routing tokens for 20 diverse tasks, covering image, video and 3D domains. Note that ``RVOS'' denotes referring video object segmentation.}}}
    \label{fig:tokens}
\end{figure*}

As illustrated in Figure~\ref{fig:tokens}, we present the task-specific routing tokens for 20 distinct computer vision tasks, spanning image, video, and 3D domains. These routing tokens play a crucial role during the training of MLLMs on \textit{OlympusInstruct}, acting as explicit indicators to guide task-specific responses. For instance, when handling a text-to-3D generation task, a sample instruction such as: \texttt{I'd appreciate it if you could design a 3D representation of an ancient library, which is a repository of books and scrolls from ancient times.''}, paired with the response: \texttt{ancient library, a repository of books and scrolls from ancient times.''}, can be augmented with the routing tokens corresponding to the text-to-3D generation task. This updated response would then appear as: \texttt{``}\verb|<3D_gen_text>|\texttt{ancient library, a repository of books and scrolls from ancient times.}\verb|</3D_gen_text>|\texttt{''}. Such augmentation ensures that the model learns to associate specific tasks with their respective routing tokens.

\begin{table*}
\centering
 \begin{tabular}{|c|c|}
\hline
\multicolumn{2}{|c|}{\textbf{Specialists for 20 individual computer tasks}} \\ \hline
\textbf{Task / Model} & \textbf{Task / Model} \\ \hline
Image Generation: Stable Diffusion XL~\citep{sdxl} & Video Generation: CogVideoX~\citep{yang2024cogvideox} \\ \hline
Image Editing: InstructPix2Pix~\citep{brooks2023instructpix2pix} & Text-to-3D Generation: LGM~\citep{tang2025lgm} \\ \hline
Video Editing: Text2Video-Zero~\citep{khachatryan2023text2video} & Image-to-3D Generation: Wonder3D~\citep{long2024wonder3d} \\ \hline
Image Deblurring: InstructIR~\citep{conde2024high} & Referring Video Object Segmentation: GLEE~\citep{wu2024general} \\ \hline
Object Detection: Co-DETR~\citep{zong2023detrs} & Image Super-Resolution: Swin2SR~\citep{conde2022swin2sr} \\ \hline
Normal Estimation: Sapiens~\citep{wu2024general} & Pose Estimation: DWPose~\citep{yang2023effective} \\ \hline
Canny Estimation: OpenCV Canny Operator  & Depth Estimation: Depth Anything V2~\citep{yang2024depth} \\ \hline
Visual Grounding: GroundingDINO~\citep{liu2023grounding} & Image Deraining: InstructIR~\citep{conde2024high} \\ \hline
Image Denoising: InstructIR~\citep{conde2024high} & Object Segmentation: SegFormer~\citep{xie2021segformer} \\ \hline
Controllable Image Generation: ControlNet~\citep{zhang2023adding} & Controllable Video Generation: Text2Video-Zero~\citep{khachatryan2023text2video} \\ 
\hline
\end{tabular}
\caption{Specified models to solve 20 different tasks.}
\label{table:sup2}
\end{table*}

By incorporating these routing tokens into the training process, the MLLMs are endowed with the ability to predict and append the appropriate tokens based on diverse user instructions during inference. This mechanism enables the invocation of the most relevant specialist models for a given task, such as \verb|<image_edit>| for image editing or \verb|<video_ref_seg>| for referring video object segmentation. The framework's modularity not only enhances task alignment but also ensures adaptability across evolving domains, facilitating the seamless integration of new tasks and specialist models in the future. This mechanism underscores the scalability and versatility of the \textit{Olympus} framework in handling complex AI tasks.

\section{Adopted Specialist Models}
\label{models}
In Table~\ref{table:sup2}, we present the specialist models selected for 20 distinct computer vision and multimodal tasks, illustrating the flexibility and adaptability of our \textit{Olympus} framework. Notably, for tasks like canny estimation, we utilize the highly efficient and widely recognized Canny operator from the OpenCV library. Similarly, for other tasks such as image generation, image editing, and text-to-3D generation, we incorporate state-of-the-art models like Stable Diffusion XL~\citep{sdxl}, InstructPix2Pix~\citep{brooks2023instructpix2pix}, and LGM~\cite{tang2025lgm}, respectively. By leveraging these specialized, task-specific models, \textit{Olympus} circumvents the need for training excessively large and cumbersome multimodal all-in-one models, instead opting for a modular and scalable approach.

\begin{table*}[h!]
\centering
\renewcommand{\arraystretch}{1.1}
\setlength{\tabcolsep}{8pt}
\begin{tabular}{|c|c|c|c|}
\hline
\multicolumn{4}{|c|}{\textbf{\# of OlympusInstruct/OlympusBench across different tasks}} \\ \hline
\textbf{Task} & \textbf{\# of samples} & \textbf{Task} & \textbf{\# of samples} \\ \hline
Image Generation & 45000 / 5000 & Video Generation & 35786 / 3976 \\ \hline
Image Editing & 34927 / 3880 & Text-to-3D Generation & 29250 / 3250 \\ \hline
Video Editing & 32227 / 3580 & Image-to-3D Generation & 10800 / 1200 \\ \hline
Image Deblurring & 6300 / 700 & Referring Video Segmentation & 21600 / 2400 \\ \hline
Object Detection & 7200 / 800 & Image Super-Resolution & 6300 / 700 \\ \hline
Normal Estimation & 6300 / 700 & Pose Estimation & 6300 / 700 \\ \hline
Canny Estimation & 6300 / 700 & Depth Estimation & 6300 / 700 \\ \hline
Visual Grounding & 23040 / 2560 & Image Denoising & 7574 / 841 \\ \hline
Image Deraining & 7650 / 850 & Object Segmentation & 7200 / 800 \\ \hline
Pose-to-Image (CIG) & 5946 / 658 & Canny-to-Image (CIG) & 5950 / 658  \\ \hline
Normal-to-Image (CIG) & 5896 / 658 & Scribble-to-Image (CIG)& 5949 / 658 \\ \hline
Segmentation-to-Image (CIG) & 5926 / 658 &  Depth-to-Image (CIG) & 5923 / 658 \\ \hline 
Pose-to-Video (CVG) & 7650 / 850 &  Canny-to-Video (CVG) & 7650 / 850 \\ \hline
Normal-to-Video (CVG) & 7650 / 850 & Scribble-to-Video (CVG) & 7650 / 850 \\ \hline
Segmentation-to-Video (CVG) & 7650 / 850 & Depth-to-Video (CVG) & 7650 / 850  \\ \hline
Chain-of-Action & 64800 / 7200 & Total & 446344 / 49585  \\  \hline
\end{tabular}
\caption{The number of collected samples for each task on \textit{OlympusInstruct}/\textit{OlympusBench}. ``CIG'' and ``CVG'' denotes controllable image generation and controllable video generation, respectively.}
\label{table:sup3}
\end{table*}

\begin{table*}[!t]
  \centering
  \footnotesize
  \resizebox{0.95\linewidth}{!}{
      \begin{tabular}{c|ccccccccccc}
        \textbf{MLLM} & \textbf{VQAv2} & \textbf{GQA} & \textbf{VisWiz} & \textbf{SQA$^{\text{I}}$} & \textbf{VQA$^{\text{T}}$} & \textbf{MME-P} & \textbf{MME-C} & \textbf{MMB} & \textbf{MM-Vet} & \textbf{POPE} & \textbf{MMMU} \\
        \midrule
        Mipha-3B &  81.3 & 63.9 & 45.7 & 70.9 & 56.6 & 1488.9 & 295.0 & 69.7 & 32.1 & 86.7 & 32.5\\
        \rowcolor{LightCyan} \textbf{Olympus-3B} & 80.5 & 63.9 & 48.2 & 70.7 & 53.4 & 1520.7 & 283.2 & 71.2 & 33.8 & 86.6 & 32.8 \\ \hline
        LLava-7B  &  78.5 & 62.0 & 50.0 & 66.8 & 58.2 & 1510.7 & 316.1 & 64.3 & 30.5 & 85.9 & 32.0\\
        \rowcolor{LightCyan} \textbf{Olympus-7B} & 78.3 & 61.9 &52.3 & 67.5  & 57.9  & 1460.2  & 328.9 & 65.6 & 32.0 & 85.4& 32.1 \\ \hline
        LLava-13B & 80.0 & 63.3 & 53.6 & 71.6 & 61.3 & 1531.3 & 295.4 & 67.7 & 36.1 & 85.9 & 33.6 \\
        \rowcolor{LightCyan} \textbf{Olympus-13B} & 79.8 & 63.1  & 56.1 & 71.9 & 61.0  & 1502.3  & 335.2  & 68.9  & 36.8 & 85.6 & 33.7 \\ \hline
      \end{tabular}
  }
    \caption{The ablation study of adopting different MLLMs across 11 benchmarks.}
  \label{sup_table1}
\end{table*}

A key strength of the \textit{Olympus} framework lies in its ability to seamlessly integrate superior models as they become available. For instance, advanced models like GroundingDINO~\citep{liu2023grounding} for visual grounding or Wonder3D~\citep{long2024wonder3d} for image-to-3D generation can be directly adopted to replace existing specialists, ensuring up-to-date performance across tasks. This modularity enables \textit{Olympus} to remain efficient and user-oriented, adapting to specific requirements without the overhead of comprehensive retraining. By selectively incorporating these external models, \textit{Olympus} provides a practical and scalable solution to address diverse and evolving user needs in various AI applications.

\section{Dataset Statistic}
\label{dataset}
We provide a comprehensive breakdown of the statistics for the \textit{OlympusInstruct} and \textit{OlympusBench} datasets in Table~\ref{table:sup3}. Specifically, the number of instruction-response pair samples is listed for each task. For example, we collected 45,000 and 5,000 samples for text-guided image generation on \textit{OlympusInstruct} and \textit{OlympusBench}, respectively. Similarly, video generation tasks include 35,786 and 3,976 samples, reflecting the diversity and scale of these datasets. Across various image and video processing tasks, such as editing, segmentation, and deblurring, we maintained a balanced distribution to ensure comprehensive task coverage.

For controllable image generation (CIG) and controllable video generation (CVG), we collected balanced samples across six specific conditions: pose, canny, normal, scribble, segmentation, and depth. Each condition is represented with approximately equal proportions to ensure robust model performance under diverse constraints. For example, pose-to-image generation includes 5,946 samples in \textit{OlympusInstruct} and 658 in \textit{OlympusBench}, while pose-to-video generation contains 7,650 and 850 samples, respectively.

In total, we collected 446,344 examples for \textit{OlympusInstruct} to serve as a comprehensive training dataset and 49,585 examples for \textit{OlympusBench} to support rigorous evaluation.

\textbf{Chain-of-Action Samples.} To simulate complex, sequential tasks, we curated instruction-response pairs for chain-of-action scenarios. Here, $N$ random tasks (ranging from 2 to 5) were selected, and one sample from each task was combined to construct multi-step instruction-response pairs. This approach resulted in 64,800 and 7,200 samples for chain-of-action tasks in \textit{OlympusInstruct} and \textit{OlympusBench}, respectively. These chain-of-action samples are designed to assess the model's ability to handle multi-step and sequential tasks within one user instruction effectively, showcasing the versatility and adaptability of the \textit{Olympus} framework.

\begin{table}[!t]
  \centering
  \footnotesize
  \resizebox{\linewidth}{!}{
      \begin{tabular}{l|cccc}
           \textbf{Method}  & \textbf{Acc} $\uparrow$ & \textbf{Pre} $\uparrow$ & \textbf{Recall} $\uparrow$ & \textbf{F1} $\uparrow$ \\
        \midrule
        HuggingGPT (GPT-4o mini) & 70.14 & 76.51 & 72.14 & 75.46 \\
        HuggingGPT (GPT-4o) & 81.35 & 85.54 & 81.55 & 83.56 \\
        \rowcolor{LightCyan} \textbf{Olympus-3B} & 94.75 & 95.80 & 94.75 & 95.77 \\
        \rowcolor{LightCyan} \textbf{Olympus-7B} &95.63  & 96.71 & 95.63 & 96.62 \\
        \rowcolor{LightCyan} \textbf{Olympus-13B} & 96.71& 97.65 & 96.71 & 96.74 
      \end{tabular}
  }
  \caption{Evaluation results on \textit{\textit{Olympus}Bench} under the single-task setting. Metrics include accuracy (\%), precision (\%), recall (\%), and F1-score (\%).}
  \label{sup_table2}
\end{table}

\section{Ablation Study}
\label{ablation}

In this section, we conduct more ablation study experiments to provide deeper insight into the effect of varying MLLMs and training epochs for \textit{Olympus}. 

\paragraph{The Impact of Adopting Different MLLMs. } To demonstrate the generality and robustness of the proposed Olympus framework, we utilize the LLava-7B and LLava-13B models as baselines, alongside Mipha-3B. These models are fine-tuned under our framework to produce Olympus-3B, Olympus-7B, and Olympus-13B, respectively. The results, summarized in Table~\ref{sup_table1}, indicate that our Olympus models achieve performance on par with or superior to their original baselines across several benchmarks. For instance, Olympus-7B demonstrates significant improvements when compared to LLava-7B, achieving higher performance on VisWiz (+2.3\%), MME-C (+12.8), MM-Vet (+1.5\%) and MMB (+1.3\%), with only a slight drop in MME-P and minimal reductions across other benchmarks. These results highlight the effectiveness of our framework in adapting existing models across multimodal tasks. Interestingly, Olympus-3B outperforms its larger counterparts, Olympus-7B and Olympus-13B, on specific benchmarks. This can be attributed to the higher input image resolution and a more robust visual encoder inherited from Mipha-3B, emphasizing the role of architectural design and input quality in determining performance.

\begin{table}[!t]
  \centering
  \footnotesize
  \resizebox{\linewidth}{!}{
      \begin{tabular}{l|cccc}
           \textbf{Method}  & \textbf{ED} $\downarrow$ & \textbf{Pre} $\uparrow$ & \textbf{Recall} $\uparrow$ & \textbf{F1} $\uparrow$ \\
        \midrule
        HuggingGPT (GPT-4o mini) & 0.45 & 65.14 & 48.51 & 53.14 \\
        HuggingGPT (GPT-4o) & 0.35 & 75.03 & 60.23 & 61.25 \\
        \rowcolor{LightCyan} \textbf{Olympus-3B} & 0.18 & 91.82 & 92.75 & 91.98 \\
        \rowcolor{LightCyan} \textbf{Olympus-7B} & 0.14 & 93.11  & 93.89  & 93.15 \\
        \rowcolor{LightCyan} \textbf{Olympus-13B} & 0.12 & 94.54 & 95.02 & 94.58 
      \end{tabular}
  }
  \caption{Evaluation results on \textit{\textit{Olympus}Bench} under the chain-of-action setting. ED represents edit distance.}
  \label{sup_table3}
\end{table}

\begin{table}[t]
  \centering
  \footnotesize
  \resizebox{0.75\linewidth}{!}{
      \begin{tabular}{l|c}
           \textbf{Method} & \textbf{Success Rate} $\uparrow$   \\
        \midrule
        HuggingGPT (GPT-4o mini) & 65.8 \\
        HuggingGPT (GPT-4o) & 75.2\\
        \rowcolor{LightCyan} \textbf{Olympus$^{*}$ (Ours)} & 80.1 \\
        \rowcolor{LightCyan} \textbf{Olympus (Ours)} & 86.5 \\
      \end{tabular}
    }
  \caption{Human evaluation of different methods, we display the results of success rate (\%). Olympus-3B is selected for evaluation, and Olympus$^{*}$ denotes training the model with coarse responses.}

\label{table:human}
\end{table}

\begin{table}[t]
  \centering
  \footnotesize
  \resizebox{1\linewidth}{!}{
      \begin{tabular}{cccc|c}
           \textbf{Example Pairs} & \textbf{Prefixes} & \textbf{Phrases} & \textbf{Complexities} & \textbf{Success Rate} $\uparrow$   \\
        \midrule
         \textcolor{red}{\xmark} & \textcolor{red}{\xmark} & \textcolor{red}{\xmark} & \textcolor{red}{\xmark} & 44.3 \\
          \textcolor{ForestGreen}{\cmark} & \textcolor{red}{\xmark} & \textcolor{red}{\xmark} & \textcolor{red}{\xmark} & 61.8  \\
         \textcolor{ForestGreen}{\cmark} & \textcolor{ForestGreen}{\cmark} & \textcolor{red}{\xmark}& \textcolor{red}{\xmark} & 68.2 \\
         \textcolor{ForestGreen}{\cmark} & \textcolor{ForestGreen}{\cmark} & \textcolor{ForestGreen}{\cmark} & \textcolor{red}{\xmark} & 76.4 \\
         \rowcolor{LightCyan} \textcolor{ForestGreen}{\cmark} & \textcolor{ForestGreen}{\cmark} & \textcolor{ForestGreen}{\cmark} & \textcolor{ForestGreen}{\cmark} & 86.5 \\
      \end{tabular}
    }
  \caption{Ablation study of different prompt components. ``Complexities" and ``Example Pairs" denote defined user instruction complexities and given user instruction-response pairs. }
\label{table:component}
\end{table}

\begin{table*}[t]
  \centering
  \footnotesize
  \begin{tabular}{cc}
    \begin{subtable}[t]{0.26\linewidth}
      \centering
      \resizebox{\linewidth}{!}{
        \begin{tabular}{c|c}
          \textbf{\# of prefixes} & \textbf{Success Rate} $\uparrow$ \\
          \midrule
          0 & 80.9 \\
          3 & 84.0 \\
          \rowcolor{LightCyan} 7 & 86.5 \\
          \rowcolor{white} 10 & 86.1 \\
        \end{tabular}
      }
      \caption{different numbers of prefixes.}
    \end{subtable}\hspace{8mm}
    \begin{subtable}[t]{0.26\linewidth}
      \centering
      \resizebox{\linewidth}{!}{
        \begin{tabular}{c|c}
          \textbf{\# of phrases} & \textbf{Success Rate} $\uparrow$ \\
          \midrule
          0 & 78.7 \\
          5 & 83.2 \\
          \rowcolor{LightCyan} 12 & 86.5 \\
          \rowcolor{white} 16 & 86.4 \\
        \end{tabular}
      }
      \caption{different numbers of phrases}
    \end{subtable}\hspace{8mm}
    \begin{subtable}[t]{0.3\linewidth}
      \centering
      \resizebox{\linewidth}{!}{
        \begin{tabular}{c|c}
          \textbf{\# of example pairs} & \textbf{Success Rate} $\uparrow$ \\
          \midrule
          0 & 69.6 \\
          \rowcolor{white} 3 & 80.3 \\
          \rowcolor{LightCyan} 9 & 86.5 \\
          \rowcolor{white} 12 & 86.3 \\
        \end{tabular}
      }
      \caption{different numbers of example pairs}
    \end{subtable}
  \end{tabular}
  
  \caption{Ablation study of varying numbers of prefixes, phrases and user instruction-response example pairs in task-specific prompts.}
   \label{table:ablation}
  \vspace{-2mm}
\end{table*}

Furthermore, as detailed in Tables~\ref{sup_table2} and~\ref{sup_table3}, Olympus-3B, Olympus-7B, and Olympus-13B achieve superior routing performance on \textit{OlympusBench} compared to HuggingGPT~\cite{shen2024hugginggpt}. In the single-task setting, Olympus models consistently exhibit higher accuracy, precision, recall, and F1 scores, with Olympus-13B achieving the best overall performance (e.g., 96.71\% accuracy and 96.74\% F1 score). Similarly, in the chain-of-action setting, Olympus models significantly reduce edit distance (ED) while maintaining high precision, recall, and F1 scores, with Olympus-13B again leading (e.g., 0.12 ED and 94.58\% F1 score). These results underscore the versatility of Olympus across diverse models and parameter scales, cementing its potential as a powerful framework for task routing.

\paragraph{The Effect of Different Response Designs.} By default, we prompt GPT-4o~\citep{hurst2024gpt} to generate concise and practical responses for each user instruction. To investigate the influence of different response designs, we also evaluate the performance of coarse responses, where the responses directly replicate the user instructions without simplification.  The trained model using coarse responses is named Olympus$^{*}$. For comparison, we report results on 200 collected human-generated instructions, as shown in Table~\ref{table:human}. The evaluation metric used is the success rate, which measures whether the specialized models generate outputs that fully satisfy user requests. As shown in Table~\ref{table:human}, Olympus$^{*}$, which adopts coarse responses, achieves a success rate of 80.1\%, while Olympus, utilizing concise and practical responses, achieves a significantly higher success rate of 86.5\%. This improvement highlights the advantage of practical responses, which are concise, direct, and focused on effectively addressing user requirements. In contrast, coarse responses, although diverse in styles, tones, and complexities, often introduce ambiguities or redundancies that can hinder the model's ability to align with user intent.

In addition, HuggingGPT (GPT-4o mini and GPT-4o) exhibits lower success rates, further underscoring the advantage of our approach for improved task routing performance.

\paragraph{The Influence of Individual Prompt Components.} To better understand the contribution of various components in task-specific prompts used for generating instruction-response pairs with GPT-4o, we conducted a detailed ablation study. The results, presented in Table~\ref{table:component}, highlight the incremental impact of including key elements such as example pairs, prefixes, phrases, varying complexity levels. Starting with a baseline configuration that excludes all these components, we observe a success rate of 44.3\%. Incorporating each individual component progressively improves performance, with the inclusion of user instruction-response example pairs alone yielding an 17.5\% gain, reaching 61.8\%. Adding prefixes further elevates the success rate to 68.2\%, while introducing phrases improves it to 76.4\%. Finally, the inclusion of complexities, alongside all the aforementioned components, results in a significant improvement, achieving the highest success rate of 86.5\%. These findings underscore the synergistic role of these components in enhancing the quality and effectiveness of task-specific prompt designs.

\paragraph{Varying Numbers of Prefixes, Phrases and Example Pairs.} In Table~\ref{table:ablation}, we present an ablation study exploring the impact of varying the numbers of prefixes and phrases in task-specific prompts by using the success rate as the evaluation metric. For prefixes, as shown in Table~\ref{table:ablation} (a), increasing the number of prefixes from 0 to 7 results in a steady improvement in success rate, with the best performance achieved at 7 prefixes (86.5\%). However, adding more prefixes beyond this point (e.g., 10) slightly reduces performance (86.1\%), indicating a diminishing return or possible overfitting when too many prefixes are used. 

Table~\ref{table:ablation} (b) analyzes the effect of using different numbers of phrases. The success rate increases significantly from 0 to 12 phrases, achieving a peak performance of 86.5\%, comparable to the optimal prefix setting. However, increasing the number of phrases to 16 results in a marginal decrease (86.4\%), again suggesting that excessive phrases may not necessarily improve performance further.

\begin{figure*}[!t]
    \centering
    \includegraphics[width=1\linewidth]{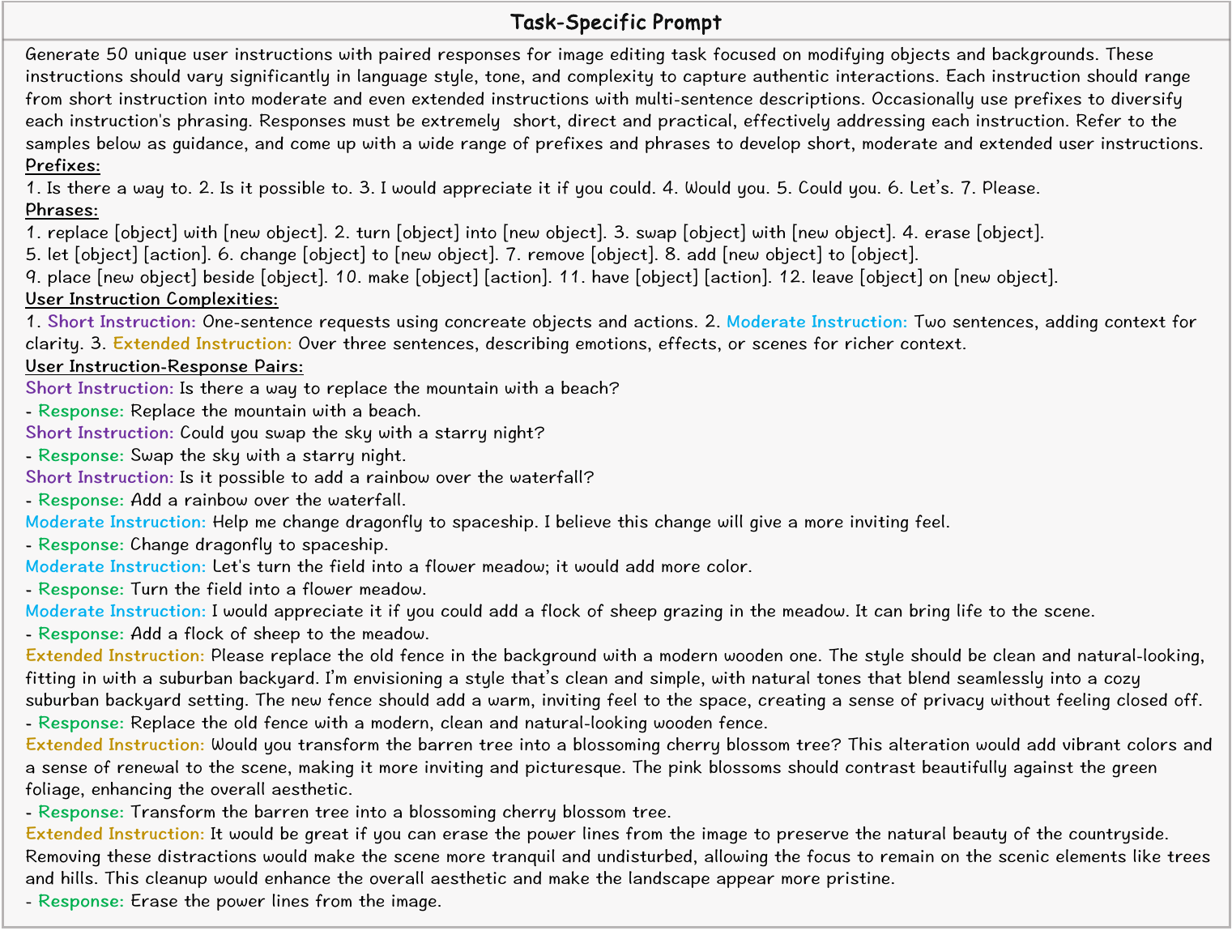}
    \caption{\small{\textcolor{black}{The final prompt used to generate user instruction-response pairs for image editing in our experiments.}}}
    \vspace{-1mm}
    \label{fig:prompt}
\end{figure*}

Similarly, Table~\ref{table:ablation} (c) examines the role of different numbers of instruction-response example pairs. Note that for different settings, the numbers of short, moderate, and extended examples remain consistent. For instance, if the number of sample pairs equals 3, there is 1 short, 1 moderate, and 1 extended example pair. The results reveal that introducing example pairs substantially boosts the success rate, from 69.6\% with no pairs to 86.5\% with 9 pairs. Interestingly, increasing the number of pairs to 12 leads to a minor performance drop (86.3\%), which may indicate a saturation effect where additional examples no longer provide significant benefit. The final prompt with 9 example pairs to generate user instruction-response pairs for the image editing task has been displayed in Figure~\ref{fig:prompt}.

These findings highlight the need for carefully calibrated choices of prefixes, phrases, and example pairs to achieve optimal task-specific prompt design, balancing informativeness and efficiency without overcomplicating the inputs.

\begin{table}[t]
  \centering
  \footnotesize
  \resizebox{0.6\linewidth}{!}{
      \begin{tabular}{c|c}
           \textbf{Complexities} & \textbf{Success Rate} $\uparrow$  \\
        \midrule
        \textcolor{red}{\xmark} & 76.8 \\
        S & 77.1 \\
        \rowcolor{white} S+M & 83.1 \\
        \rowcolor{LightCyan} S+M+E & 86.5 \\
      \end{tabular}
    }
  \caption{The ablation study of different complexity levels defined in the task-specific prompts. ``S'', ``M'' and ``E'' represent short, moderate and extended complexities respectively. Note that the term \textcolor{red}{\xmark} in the ``Complexities'' column indicates the absence of any complexity definitions in the task-specific prompts.}
  \label{tab:complexity}
\end{table}

\paragraph{The Complexities of User Instructions.} As shown in Table~\ref{tab:complexity}, we analyze the impact of varying instruction complexities in task-specific prompts on the model's success rate. By default, without any specific complexity definitions, the baseline achieves a success rate of 76.8\%. Introducing the definition of short (S) instruction complexity yields a negligible improvement, raising the success rate to 77.1\%. When moderate (M) complexities are added alongside short ones (S+M), the success rate rises significantly to 83.1\%, demonstrating the benefits of incorporating greater linguistic diversity and intermediate complexity. Finally, further defining "extended" (E) instruction complexity leads to the model achieving its highest performance, with short, moderate, and extended instructions (S+M+E) resulting in a success rate of 86.5\%. This progression underscores the critical role of instruction complexity in enhancing the model's ability to effectively interpret and execute user requests.

\section{More Results}
\label{visualization}
\begin{figure*}[!t]
    \centering
    \includegraphics[width=1\linewidth]{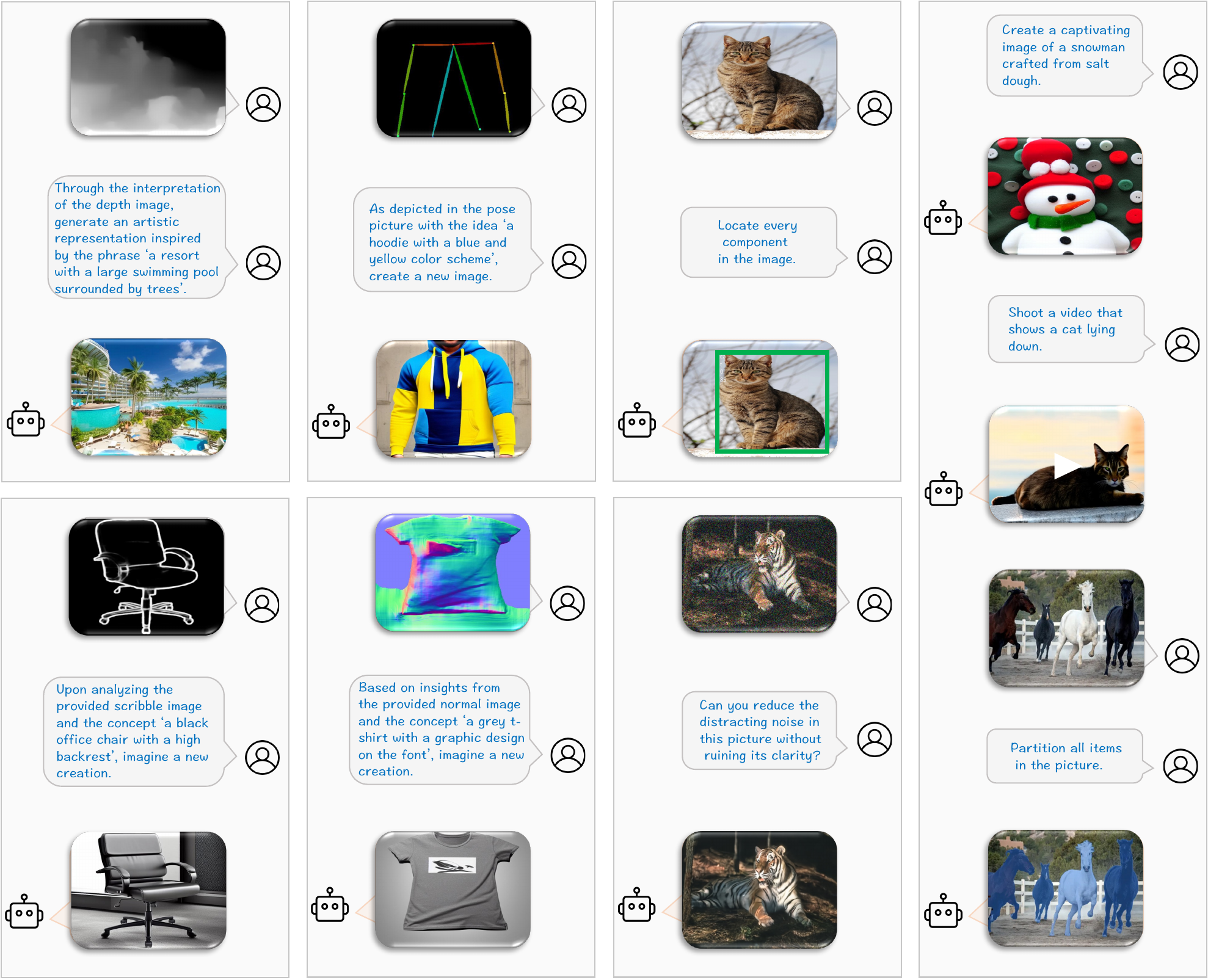}
    \caption{\small{\textcolor{black}{Diverse applications of \textit{Olympus}. The first and second columns denote controllable image generation conditioning on depth, pose, scribble and normal. The third column represent the applications of object detection and image deblurring. The last column represent the applications of text-guided image and video generation, and object segmention.}}}
    \vspace{-1mm}
    \label{fig:applications}
\end{figure*}

We present additional application results in Figure~\ref{fig:applications}, showcasing the versatility of our approach across a wide range of tasks. The first and second columns demonstrate controllable image generation conditioned on diverse inputs such as depth maps, poses, scribbles, and surface normals, illustrating the model's flexibility to adapt to varying conditions. The third column highlights advanced applications like object detection and image deblurring, emphasizing the framework's utility in real-world image processing. Lastly, the fourth column explores cutting-edge tasks, including text-guided image and video generation and object segmentation, underscoring the model’s potential for multimodal and domain-specific applications.

\begin{figure*}[!t]
    \centering
    \includegraphics[width=1\linewidth]{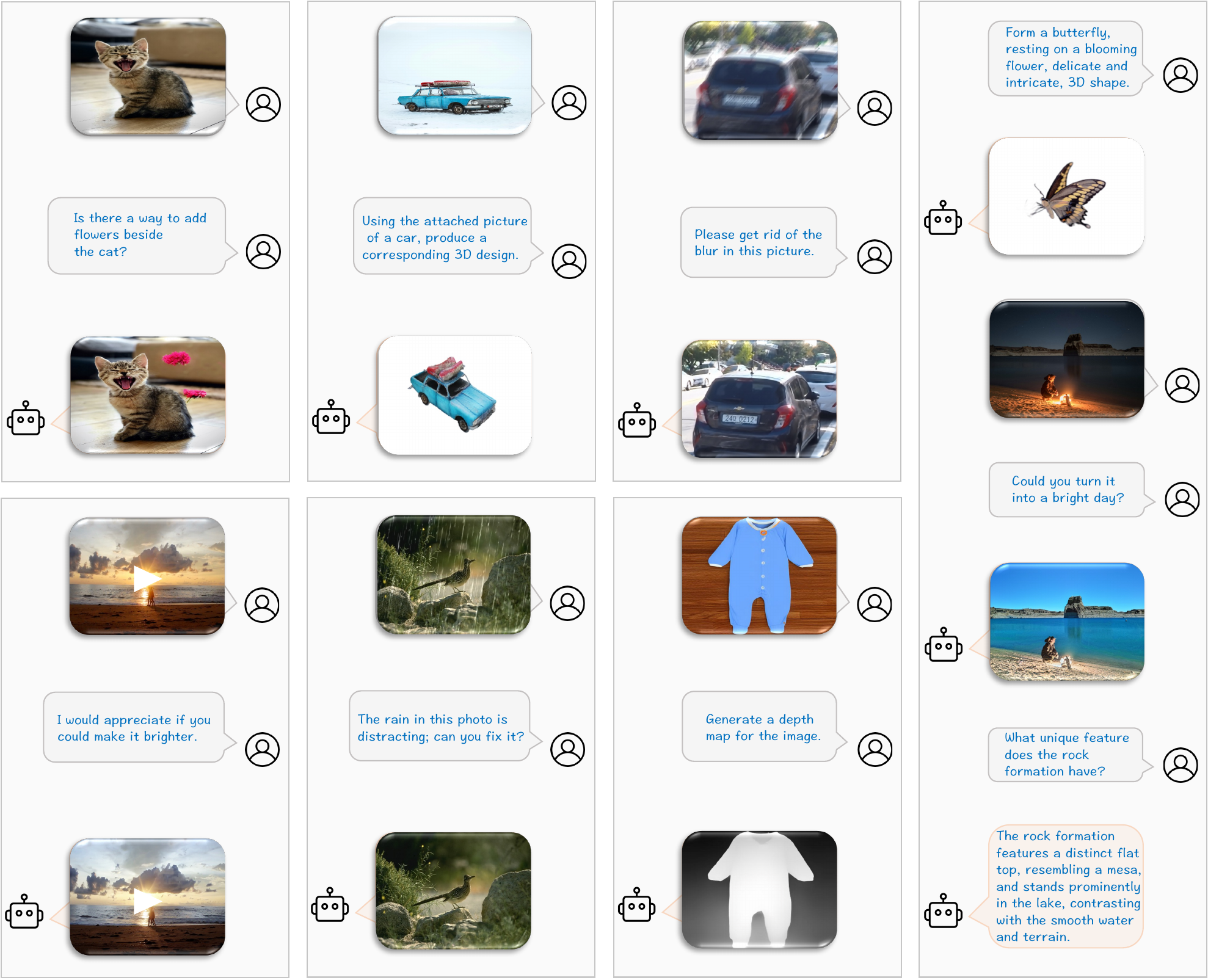}
    \caption{\small{\textcolor{black}{Diverse applications of \textit{Olympus}. The first column represents image and video editing, the second column contains the examples of image-to-3D generation and image deraining, the third column denotes image deblurring and depth estimation, and the final column displays the applications of text-to-3D generation, image editing and visual question answering (VQA).}}}
    \vspace{-1mm}
    \label{fig:applications2}
\end{figure*}

In Figure~\ref{fig:applications2}, the first column illustrates image and video editing tasks, such as adding flowers beside a cat or brightening a sky. The second column highlights image-to-3D generation and image deraining, including transforming a 2D car image into a 3D model and removing rain effects to improve clarity. The third column focuses on image deblurring, such as removing motion blur from a car photo, and depth estimation, demonstrated by generating depth maps from images. The fourth column features diverse applications, including text-to-3D generation (e.g., creating intricate 3D shapes from a detailed prompt for "a butterfly"), editing environmental lighting (e.g., converting night to day), and visual question answering (VQA), where image content questions, such as identifying unique rock formation features, are answered with precision.

These results demonstrate the strong generality and scalability of \textit{Olympus}, highlighting its adaptability and effectiveness across a broad and growing set of tasks.

\clearpage
\clearpage
\clearpage


\end{document}